\documentclass[lettersize,journal]{IEEEtran}
\usepackage{amsmath,amsfonts}
\usepackage{algorithmic}
\usepackage{algorithm}
\usepackage{array}
\usepackage[caption=false,font=normalsize,labelfont=sf,textfont=sf]{subfig}
\usepackage{textcomp}
\usepackage{stfloats}
\usepackage{url}
\usepackage{verbatim}
\usepackage{graphicx}
\usepackage{cite}
\usepackage{cases}

\usepackage{xcolor,colortbl}
\definecolor{LightCyan}{rgb}{0.88,1,1}
\definecolor{LightRed}{rgb}{1,0.92,0.92}

\usepackage{color}

\newcommand{\bx}{\mathbf{x}}
\newcommand{\bq}{\mathbf{q}}
\newcommand{\bphi}{\boldsymbol{\phi}}

\newcommand{\bomega}{\boldsymbol{\omega}}

\newcommand{\bnu}{\boldsymbol{\nu}}
\newcommand{\balpha}{\boldsymbol{\alpha}}
\newcommand{\bgamma}{\boldsymbol{\gamma}}
\newcommand{\boeta}{\boldsymbol{\eta}}
\newcommand{\bbeta}{\boldsymbol{\beta}}
\newcommand{\bzeta}{\boldsymbol{\zeta}}

\begin{document}

\title{Grasp Failure Constraints for Fast and Reliable Pick-and-Place Using Multi-Suction-Cup Grippers}

\author{Jee-eun Lee$^{1}$, Robert Sun$^{2}$, Andrew Bylard$^{2, *}$, and Luis Sentis$^{1, *}$
\thanks{This work was supported by Dexterity, Inc.}
\thanks{* Corresponding authors equally contributed }
\thanks{$^{1}$ J. Lee and L. Sentis are with The University of Texas at Austin, Austin, TX, USA, 
{\tt\small \{jelee, lsentis\}@utexas.edu}}%
\thanks{$^{2}$ A. Bylard, and R. Sun are with Dexterity, Inc., Redwood City, CA, USA, 
{\tt\small \{andrew.bylard, robert\}@dexterity.ai }}%
\thanks{Manuscript received April 19, 2021; revised August 16, 2021.}}

\markboth{Journal of \LaTeX\ Class Files,~Vol.~14, No.~8, August~2021}%
{Shell \MakeLowercase{\textit{et al.}}: A Sample Article Using IEEEtran.cls for IEEE Journals}

\IEEEpubid{0000--0000/00\$00.00~\copyright~2021 IEEE}

\maketitle

\begin{abstract}
Multi-suction-cup grippers are frequently employed to perform pick-and-place robotic tasks, especially in industrial settings where grasping a wide range of light to heavy objects in limited amounts of time is a common requirement. However, most existing works focus on using one or two suction cups to grasp only irregularly shaped but light objects. There is a lack of research on robust manipulation of heavy objects using larger arrays of suction cups, which introduces challenges in modeling and predicting grasp failure. This paper presents a general approach to modeling grasp strength in multi-suction-cup grippers, introducing new constraints usable for trajectory planning and optimization to achieve fast and reliable pick-and-place maneuvers. The primary modeling challenge is the accurate prediction of the distribution of loads at each suction cup while grasping objects. To solve for this load distribution, we find minimum spring potential energy configurations through a simple quadratic program. This results in a computationally efficient analytical solution that can be integrated to formulate grasp failure constraints in time-optimal trajectory planning. Finally, we present experimental results to validate the efficiency and accuracy of the proposed model.
\end{abstract}

\def\abstractname{Note to Practitioners}
\begin{abstract}
Pick-and-place tasks are common in logistics. However, handling heavy objects can cause musculoskeletal disorders, and workplaces with extreme temperatures exceeding 40°C are often unsuitable for human workers, necessitating robotic automation. While maximizing robot speed is crucial for productivity, it also increases the risk of grasp failures, which can potentially damage fragile or deformable objects during handling. This paper addresses these challenges by introducing new grasping failure constraints tailored for a vacuum gripper with multiple suction cups. Integrating these constraints into time-optimal trajectory planning algorithms enables robots to safely and efficiently handle a variety of objects, thereby enhancing overall productivity.
\end{abstract}


\section{Introduction}
\IEEEPARstart{W}{ithin} the domain of industrial automation, the quest to increase speed and reliability in object manipulation remains a core engineering challenge. In this pursuit, vacuum grippers have proven to be valuable tools, offering broad capabilities for handling diverse objects across manufacturing and logistics applications~\cite{reddy2013review, jaiswal2017vacuum}. Among various types of vacuum grippers, including suction plates~\cite{wirth2020suctionplate}, foam grippers, and sack grippers specialized for bag handling, those with multi-suction-cup have seen widespread use due to their versatility and strong grip \cite{correll2016analysis, mykhailyshyn2022gripping,  zhang2020state, jo2024suction}. However, multi-suction-cup grasp failure is a complex, multivariate phenomenon that is challenging to model. Predicting failure requires determining the load distribution on each suction cup, which can be most accurately computed using detailed system identification and Finite Element Analysis (FEA) \cite{seretse2023material, joymungul2021gripe}. 
However, due to its high computation requirements, FEA is at present unsuitable for direct integration into real-time motion planning algorithms where grasp modeling can inform trajectory planning. As a result, during motion generation, direct grasp failure constraints are often replaced with heuristics, such as acceleration and deceleration limits, chosen through trial and error. 

Extensive efforts have been devoted to formulating grasp constraints for trajectory planning. Several recent studies, in particular, have focused on finding the optimal grasp configurations for objects with irregular shapes under the assumption of quasi-static movement. For example, Grasp-Optimized Motion Planning (GOMP) \cite{ichnowski2020gomp} and Deep-Jerk GOMP (DJ-GOMP) \cite{ichnowski2020deep} compute time-optimal pick-and-place motions by solving a sequential quadratic program. They leverage learned policies to determine the optimized grasp pose for parallel-jaw grippers. However, these approaches do not guarantee a secure grasp during rapid motions due to their quasi-static assumption. GOMP-fit \cite{ichnowski2022gomp} proposes limiting the acceleration
of the objects being transported by adding end-effector constraints, as described in \cite{lynch1996dynamic, lynch1999dynamic}. However, this method relies on heuristically predefined thresholds for the constraints, which can lead to unnecessarily conservative and slower motions. 

\IEEEpubidadjcol
While there is extensive research on handling objects with irregular shapes, fast pick-and-place of heavy objects has received less attention, even though it is a common requirement in industry. Such objects are often handled using multi-suction-cup grippers, which deserve special attention due to their complex dynamics. The work in \cite{zhu2002development} presents a theoretical approach to determine the payload capacity of suction cups for climbing robots, though it oversimplifies the problem by assuming uniform load distribution between cups. More comprehensive models as presented in \cite{mantriota2007optimal, mantriota2007theoretical}, take into account 3D force factors, but these models are limited to specific arrangements of grippers' suction cups. To address complex dynamics in suction-cup grasping, GOMP-st \cite{avigal2022gomp} uses learning-based methods to formulate grasp constraints, integrating them into conventional solvers using differentiable activation functions as described in \cite{de2018learning, fajemisin2024optimization}.
Indeed, learning-based methods may excel in predicting complex and unstructured interactions, such as grasping irregularly shaped objects~\cite{kleeberger2020survey,mahler2018dex}, given their ability to handle the multitude of factors that are often overlooked in traditional theoretical models.
However, the performance of learned models heavily relies on the quality and diversity of the training datasets, making them vulnerable to new or unseen scenarios. In contrast, methods based on physical principles offer a more universally applicable approach. While some data collection for system identification may still be necessary to enhance overall algorithm accuracy, the quantity and complexity of the required data are considerably lower, presenting a significant advantage for industrial applications.

The research most similar to ours is from study \cite{pham2019critically}, which aims to generate fast pick-and-place motions using time-optimal trajectory planning for a gripper with a single suction cup. It leverages a weak contact stability formulation ~\cite{caron2015leveraging}, widely used in multi-contact trajectory optimization~\cite{carpentier2018multicontact, fernbach2020c}. This approach checks for the existence of a feasible solution that satisfies suction grasp constraints and wrench equations using the double description method (DD method). However, since the problem is underdetermined (as explained in Section \ref{sec:grasp_failure}), there is an infinite set of solutions satisfying wrench balance. Many of these solutions are far from the true load distribution, and therefore ineffective for predicting grasp failure. Instead, we aim at incorporating additional physical principles to select load distribution solutions that are closer to real world interactions. 

In this paper, we introduce an approach to find a more realistic solution for predicting grasp failures in multi-suction-cup vacuum grippers. Based on the principle of minimum spring potential energy, we formulate a load distribution problem as a quadratic program (QP). This allows us to compute the loads distributed on each suction cup with an analytical solution, similar to \cite{lee2022adaptive}. This load distribution model can then be used to formulate grasp failure constraints throughout the object's trajectory. Finally, we demonstrate that this approach can be successfully integrated into time-optimal trajectory planning to generate fast and robust pick-and-place motions for robots manipulating heavy payloads with multi-suction-cup grippers.

\subsection{Contributions}
\begin{itemize}
    \item We propose the first analytical grasp failure model generalized for any configuration of vacuum grippers with multiple suction cups, providing a solution simple enough to be integrated into time-optimal trajectory planning (TOTP).
    \item We rigorously verify the accuracy of our load distribution model and grasp failure conditions using a testbed gripper equipped with force sensors, capable of measuring load distribution across multiple suction cups.
    \item We demonstrate that integrating the proposed grasp failure constraint into TOTP algorithm on a real robot effectively reduces grasp failures while ensureing fast movements.
    This improvement is substantiated by the reduced false positive rate observed in the statistical analysis of real robot experiments.
\end{itemize}


\subsection{Nomenclature}
To ensure clarity on the notation used for vectors, transformations, and adjoint mappings across different frames, Table \ref{tab:Nomenclature} presents the conventions employed in this paper, which are mostly similar to those in \cite{lynch2017modern}.

\begin{table}[t!]
\setlength{\tabcolsep}{3pt}
\renewcommand{\arraystretch}{1.2}
\caption{Symbols and notations used in the paper\label{tab:Nomenclature}}
\vspace{-2mm}
\centering
\begin{tabular}{c p{6.4cm}}
\hline
Symbol & Definition \\
\hline
$\{a\}$ & Reference frame ``$a$" \\
$x_{1:n}$ &  Vector of $\mathbf{x}=[x_1, x_2, \cdots, x_n]^\top$\\ 
$\mathbf{x}_a$ & Vector whose reference frame is $\{a\}$ \\
$\overline{\mathbf{x}}$ & $\mathbf{x}$ in homogeneous coordinates, $\overline{\mathbf{x}}=[\mathbf{x}^\top 1]^\top$\\
$\mathbf{p}_a^b \in \mathbb{R}^3 $ & Position of frame $\{b\}$ with respect to $\{a\}$\\
$\mathbf{R}_a^b \in SO(3)$ & Orientation of $\{b\}$ expressed in $\{a\}$ or a rotation matrix to change the reference frame from $\{b\}$ to $\{a\}$ (e.g., $\mathbf{p}_a = \mathbf{R}_a^b\mathbf{p}_b$)\\ 
$\mathbf{T}_a^b \in SE(3)$ 
& Homogeneous transformation matrix that maps quantities expressed from $\{b\}$ to $\{a\}$. (e.g., $\overline{\mathbf{p}}_a = \mathbf{T}_a^b \overline{\mathbf{p}}_b$). It is also noted as $(\mathbf{R}_a^b,\mathbf{p}_a^b)=\begin{bmatrix}\mathbf{R}_a^b & \mathbf{p}_a^b \\ 0 & 1\end{bmatrix}$. \\
$[\mathbf{x}]_\times$ & Cross-product operator in skew-symmetric matrix form, meaning that $\mathbf{x}\times\mathbf{y} = [\mathbf{x}]_\times \mathbf{y}$. \\ 
& $[\mathbf{x}]_\times = \begin{bmatrix} 0 & -x_3 & x_2 \\ x_3 & 0 & -x_1 \\ -x_2 & x_1 & 0 \end{bmatrix}$ for $\mathbf{x}=(x_1, x_2, x_3)$. \\
 $\mathcal{F}=\begin{bmatrix} \mathbf{m} \\ \mathbf{f} \end{bmatrix} \in \mathbb{R}^6$ & $\mathcal{F}=(\mathbf{m}, \mathbf{f})$ represents a wrench where $\mathbf{m}$ represents a moment and $\mathbf{f}$ represents a force\\
$\mathcal{F}_a^b, \mathcal{F}_a\in \mathbb{R}^6$  & Wrench force acting at $\{b\}$ w.r.t. the frame $\{a\}$ or wrench force where the acting frame and the reference frame are both $\{a\}$. \\
 $\mathcal{V} =\begin{bmatrix} \bomega \\ \mathbf{v} \end{bmatrix}\in \mathbb{R}^6$ & $\mathcal{V}=(\bomega, \mathbf{v})$ represents a twist, where $\bomega$ is angular velocity and $\mathbf{v}$ is linear velocity. \\
 $\mathcal{V}_a^b, \mathcal{V}_a\in \mathbb{R}^6$  & Velocity of $\{b\}$ represented in $\{a\}$, or a velocity where the target frame and the reference frame are both $\{a\}$.\\
$[Ad_T]\in \mathbb{R}^{6\times6}$ & $[Ad_T]=\begin{bmatrix}
R & 0 \\ [p]_\times R & R \end{bmatrix}$ is the adjoint representation that maps twists and wrenches represented in different frames: 
$\mathcal{V}_a = [\textrm{Ad}_{\mathbf{T}_a^b}] \mathcal{V}_b , \quad \mathcal{F}_a = [\textrm{Ad}_{\mathbf{T}_b^a}]^\top \mathcal{F}_b$  \vspace{1pt} \\
\hline
\end{tabular}
\vspace{-5mm}
\end{table}

%
%

\section{Failure Constraints in Multiple-suction-cup Systems} \label{sec:grasp_failure}

For the purpose of this work, grasp failure is defined as the unintentional release of an object from the gripper's grasp. Other potential failures during grasp, such as object damage due to poor structural cohesion or items falling out of open containers, are not considered. Regarding suction-cup grippers, we distinguish two main contributing factors for grasp failure: suction loss and slippage. In this section, we establish the theoretical constraints to prevent both issues. Given that grasp failure conditions depend on the load applied to each cup in the multi-suction-cup gripper case, our formulation must account for load distribution among the cups. Accordingly, in this section, we introduce a load distribution model proposed for this work.

\subsection{Constraints to Prevent Suction Loss}
When a robot uses suction cups to hold objects, the loss of even a single suction cup during motion can significantly increase the force on the remaining suction cups, increasing the risk of losing the entire grasp. It may be possible for the remaining cups to maintain grasp after the load is redistributed, but for simplicity, we define suction loss constraints based on the following assumption:

\textit{\textbf{Assumption 1} (Suction Loss Condition)}. We assume that grasp failure due to suction loss occurs if the pulling force exerted on any suction cup exceeds its suction force.

\begin{figure}[t]
    \centering
    \includegraphics[width=0.9\linewidth]{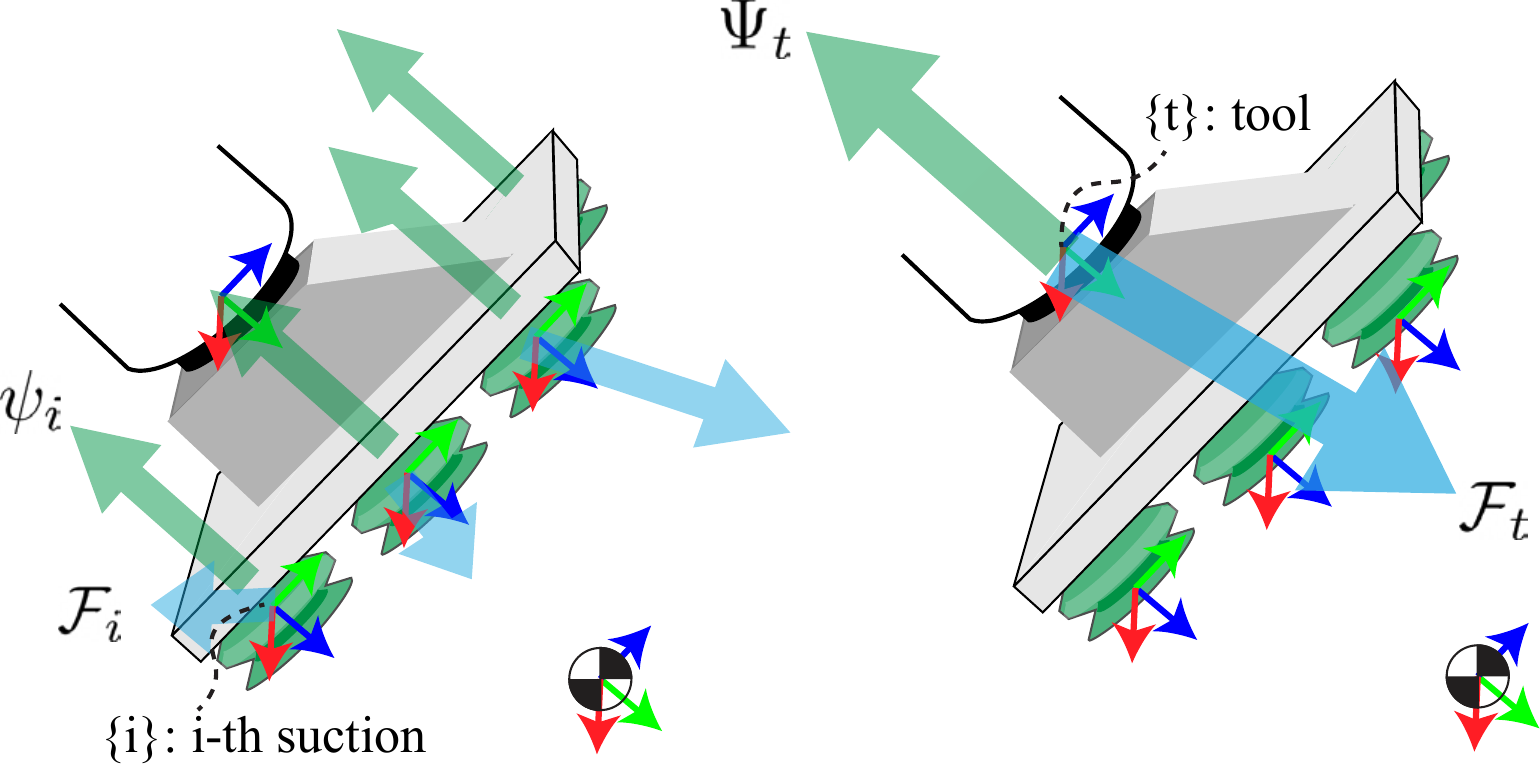}
    \vspace{-2mm}
    \caption{Illustration of the reaction forces exerted by gravity and the movement of a hypothetical object being carried by the gripper (not shown). The left figure shows the distributed reaction force (blue arrows) and the suction force (green arrows) on each suction cup. The right figure shows the sum of wrenches at the origin of the tool frame of the robot.  }
    \label{fig:force_distribution}
    \vspace{-5mm}
\end{figure}

As shown in Fig. \ref{fig:force_distribution}, let $\mathcal{F}_i = (\mathbf{m}_i, \mathbf{f}_i)\in\mathbb{R}^6$ represent the reaction wrench applied to a box at each suction cup frame $\{i\}$ origin, where each frame's $z$-direction is aligned with the normal direction of the corresponding suction cup. Given the suction force $\psi_{i}$ generated by the $i$-th suction cup, we can write the condition for not losing cup suction as
\begin{align}\label{eqn:sucloss_ineq}  
\mathbf{U}_s  \mathcal{F}_i \leq \mathbf{u}_s, 
\end{align}
where $\mathbf{U}_s$ and $\mathbf{u}_s$ can be derived from the following inequality formulated for the suction cup, considering a radius of $r_\textrm{pad}$:
\begin{equation*}
\begin{bmatrix} \ \ 0 & \ \ 0 & 0 & 0 & 0 & -1 \\
\ \ 1 & \ \ 1 & 0 & 0 & 0& -r_\textrm{pad} \\ 
\ \ 1 & -1 & 0 & 0 & 0& -r_\textrm{pad} \\ 
-1 & \ \ 1 & 0 & 0 & 0& -r_\textrm{pad} \\ 
-1 & -1 & 0 & 0 & 0& -r_\textrm{pad} 
\end{bmatrix} 
( \mathcal{F}_i +
 \begin{bmatrix} 0\\ 0 \\ 0 \\ 0 \\ 0 \\ \psi_i \end{bmatrix} ) \leq 0.
\end{equation*}
Note that the wrench condition above for secure grasp resembles the form commonly used to formulate stable contact constraints \cite{caron2015stability}. By stacking constraints for all suction cups, we can formulate the complete set of wrench conditions as
\begin{equation}\label{eqn:sucloss_ineq_full}
  \overline{\mathbf{U}}\, \overline{\mathcal{F}} \leq \overline{\mathbf{u}},
\end{equation}
where
\begin{equation*}
\overline{\mathbf{U}} = 
\begin{bmatrix}
\mathbf{U_s} & 0 &\cdots \\
 0  & \ddots & 0 \\
\vdots & 0 & \mathbf{U_s}
\end{bmatrix}, \; 
\overline{\mathcal{F}} = \begin{bmatrix}
    \mathcal{F}_{1} \\ \vdots \\ \mathcal{F}_{N_s}
\end{bmatrix}, \;
\overline{\mathbf{u}} = 
\begin{bmatrix}
\mathbf{u_s} \\ \vdots \\ \mathbf{u_s}
\end{bmatrix}.
 \end{equation*}

\subsection{Constraints to Prevent Grasp Slippage}
If loaded grasp forces exceed the friction threshold, the grasp may slip, leading to grasp failure. However, unlike suction loss constraints that are enforced on an individual cup-level basis, slippage of the entire grasp can still be prevented by the combined gripping force of the other suction cups, even if one suction cup is in a slip condition. Thus, we establish slippage constraints based on the following assumption:

\textit{\textbf{Assumption 2} (Slippage Condition)}. We assume that slippage occurs if the total force required on the suction cups to maintain grasp exceeds the friction cone inequality.

Let $\mathcal{F}_t = (\mathbf{m}_t, \mathbf{f}_t) \in \mathbb{R}^6$ represent the total wrench, computed by adding all forces applied by the suction cups to the box. In particular, we use a wrench expressed in the tool frame $\{t\}$, positioned at the center of the suction cups. Similarly, we define $\Psi_{t} \in \mathbb{R}^6$, as the sum of suction wrenches expressed in the tool frame. Finally, given the minimum polygon that includes all suction cups, let $X, Y$ be the maximum distance to the boundary of the polygon from the tool frame in the $x, y$ direction. Then, given a friction coefficient $\mu$ at the interface between each suction cup and a grasped object, we have
\begin{equation} \label{eqn:slippage_ineq}
\mathbf{U_t}  \mathcal{F}_t \leq \mathbf{u}_t,
\end{equation}
which can be derived from
\begin{align*}
\small{
\begin{bmatrix} 
\ \ 0 & \ \ 0 & \ \ 0 & \ \ 1 & \ \ 1 & -\mu \\ 
\ \ 0 & \ \ 0 & \ \ 0 & \ \ 1 & -1& -\mu \\
\ \ 0 & \ \ 0 & \ \ 0 & -1 & \ \ 1& -\mu \\
\ \ 0 & \ \ 0 & \ \ 0 & -1 & -1& -\mu  \\
\ \ \mu& \ \ \mu& -1& -Y& -X& -\mu(X+Y)  \\
\ \ \mu& -\mu& -1& -Y& \ \ X& -\mu(X+Y)  \\
-\mu& \ \ \mu& -1& \ \ Y& -X& -\mu(X+Y)  \\
-\mu& -\mu& -1& \ \ Y& \ \ X& -\mu(X+Y)  \\
\ \ \mu& \ \ \mu& -1& \ \ Y& \ \ X& -\mu(X+Y)  \\
\ \ \mu& -\mu& -1& \ \ Y& -X& -\mu(X+Y)  \\
-\mu& \ \ \mu& -1& -Y& \ \ X& -\mu(X+Y)  \\
-\mu& -\mu& -1& -Y& -X& -\mu(X+Y)  
\end{bmatrix} } & \nonumber \\
 \cdot \begin{bmatrix}
    R_i^t & 0 \\ 0 & R_i^t
\end{bmatrix}  ( \mathcal{F}_t &+ \Psi_{t} )  \leq 0, 
\end{align*}

where
\begin{align}
  \mathcal{F}_{t} &= \sum_{i \in \textrm{suction}}[\textrm{Ad}_{\mathbf{T}_i^t}]^\top \mathcal{F}_i \nonumber \\
   &= \underbrace{
\Bigg[
    [\textrm{Ad}_{\mathbf{T}_1^t}]^\top \quad \cdots \quad [\textrm{Ad}_{\mathbf{T}_{N_s}^t}]^\top
\Bigg] }_{\mathbf{A}_\textrm{g}}
\begin{bmatrix}
    \mathcal{F}_{1} \\ \vdots \\ \mathcal{F}_{N_s}
\end{bmatrix} \nonumber \\
&:= \mathbf{A}_\textrm{g}\overline{\mathcal{F}}, \label{eqn:gripper_config}\\
\Psi_{t} &= \sum_{i \in \textrm{suction}}[\textrm{Ad}_{\mathbf{T}_i^t}]^\top [0,0,0,0,0,\psi_i]^\top.
\end{align} 

\subsection{Load Distribution Model} \label{subsec:load_distribution}

\begin{figure}[t]
\centering    
\includegraphics[width=\linewidth]{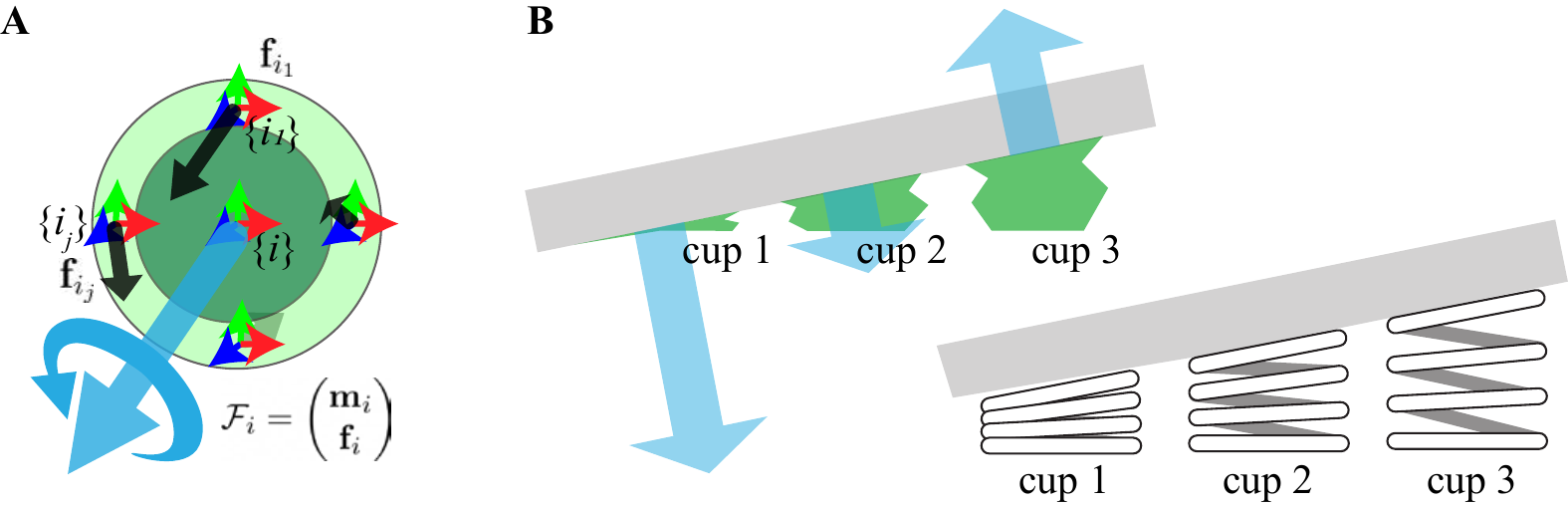}
\vspace{-3mm}
\caption{Figure A illustrates point force distribution on a single suction cup. The wrench, comprising a 3d moment and 3d force applied to the suction cup, is represented as a sum of 3d point forces distributed along the rim of the suction pad. Figure B visualizes a compressed suction cup as a spring under compression. The compressed suction cup is expected to exhibit increased stiffness, especially in the direction of compression.}
\label{fig:pointforce_compressed}
\vspace{-5mm}
\end{figure}

The total load distributed across the suction cups due to an object's movement can be determined using the equations of motion (the detailed equations are formulated in Section \ref{sec:mop}). When there are multiple suction cups on a gripper, computing the wrenches applied to each suction cup becomes an underdetermined problem. This means that there are more variables (six wrench components per cup) to be determined than there are equations available to solve for them (six equations total), as illustrated by Equation (\ref{eqn:gripper_config}). To tackle this problem, we propose an approximate force distribution model based on the following assumptions:

\textit{\textbf{Assumption 3} (Wrench exerted at a suction cup)}. We assume that a wrench exerted on a suction cup can be represented as a sum of point forces applied to the cup's ring, similarly to \cite{pham2019critically}. As depicted in Fig.~\ref{fig:pointforce_compressed}A, we consider 4 points on the ring ($\mathcal{J}=\{1,2,3,4\}$), and denote the frame attached to these points as $\{i_j\}$. We can then write:
\begin{align}
\mathcal{F}_i = \begin{bmatrix}
    \mathbf{m}_i \\ \mathbf{f}_i
\end{bmatrix} = \sum_{j\in \mathcal{J}} \begin{bmatrix}
    [\mathbf{p}_i^{i_j}]_\times \\ \mathbf{I}_{3\times 3}.
\end{bmatrix} \mathbf{f}_{i_j} \label{eqn:cup_config}
\end{align}

\textit{\textbf{Assumption 4} (Minimum potential energy model)}. We assume that the forces applied to each suction cup can be modeled as 3D spring forces, with the natural force distribution being the one that minimizes the potential energy of the springs.

\textit{\textbf{Assumption 5} (Hooke's Law)}. We assume linear suction cup spring forces with constant stiffness, represented by the equation $\mathbf{f}_{i_j} = \mathbf{K}_i \Delta \mathbf{x}_{i_j}$. For simplicity, we assume that the stiffness matrix is diagonal  $\mathbf{K}_i = \textrm{diag}([k_i^1, k_i^2, k_i^3])$, and that the stiffness is uniform across all points of the suction cup.

We can then express the minimum spring potential energy of our system as follows:
\begin{equation}
\textrm{min}~ \dfrac{1}{2} \sum_{i\in \textrm{suction}} \sum_{j\in \mathcal{J}} {\Delta\bx_{i_j}}^\top \mathbf{K}_i \Delta\bx_{i_j}.
\end{equation}

Using this approach, we can express the equation in terms of forces by substituting  $\Delta\bx_{i_j} = \mathbf{K}_i^{-1} \mathbf{f}_{i_j}$ and then represent the summation in a vector-matrix form by stacking forces into a column vector and using a block-diagonal full stiffness matrix. Then, we obtain:
\begin{align*}
\sum_{i\in \textrm{suction}} \sum_{j\in \mathcal{J}} &{\Delta\bx_{i_j}}^\top \mathbf{K}_i \Delta\bx_{i_j} \\
=& \sum_{i\in \textrm{suction}} \sum_{j\in \mathcal{J}} \mathbf{f}_{i_j}^\top {\mathbf{K}_i}^{-1} \mathbf{f}_{i_j} \\ 
=& \sum_{i\in \textrm{suction}} 
\begin{bmatrix}
    \mathbf{f}_{i_1} \\ \vdots \\ \mathbf{f}_{i_4}
\end{bmatrix}^\top
\underbrace{ \begin{bmatrix}
    {\mathbf{K}_i}^{-1} & &\\ & \ddots & \\ & & {\mathbf{K}_i}^{-1}
\end{bmatrix}}_{\mathbf{W}_i}
\underbrace{ \begin{bmatrix}
    \mathbf{f}_{i_1} \\ \vdots \\ \mathbf{f}_{i_4}
\end{bmatrix}}_{\overline{\mathbf{f}}_{i}}  \\ 
=& \begin{bmatrix}
    \overline{\mathbf{f}}_{1} \\ \vdots \\ \overline{\mathbf{f}}_{N_s}
\end{bmatrix}^\top
\underbrace{ \begin{bmatrix}
    \mathbf{W}_1 & &\\ & \ddots & \\ & & \mathbf{W}_{N_s}
\end{bmatrix} }_{\mathbf{W}}
\underbrace{ \begin{bmatrix}
    \overline{\mathbf{f}}_{1} \\ \vdots \\ \overline{\mathbf{f}}_{N_s}
\end{bmatrix} }_{\overline{\mathbf{f}}} \\ 
=& ~\overline{\mathbf{f}}^\top \mathbf{W} \overline{\mathbf{f}}.
\end{align*}


Similarly, we reformulate the constraint on the total force expressed in Equations (\ref{eqn:gripper_config}) and (\ref{eqn:cup_config}) in vector-matrix form:
\begin{align*}
\mathcal{F}_i &=
\begin{bmatrix}
    [\mathbf{p}_i^{i_1}]_\times & \cdots &
    [\mathbf{p}_i^{i_4}]_\times \\ 
     \mathbf{I}_{3\times 3} & & \mathbf{I}_{3\times 3}
\end{bmatrix} 
\begin{bmatrix}
    \mathbf{f}_{i_1} \\ \vdots \\ \mathbf{f}_{i_4}
\end{bmatrix} := \mathbf{A}_\textrm{i}\overline{\mathbf{f}}_i, \\ 
\overline{\mathcal{F}} &= \begin{bmatrix}
    \mathcal{F}_{1} \\ \vdots \\ \mathcal{F}_{N_s}
\end{bmatrix} = 
\begin{bmatrix}
    \mathbf{A}_1 & 0 & \\
    0 & \ddots & 0 \\
     & 0 & \mathbf{A}_{N_s}
\end{bmatrix}
\begin{bmatrix}
    \overline{\mathbf{f}}_1 \\ \vdots \\ \overline{\mathbf{f}}_{N_s}
\end{bmatrix} :=
\mathbf{A}_\textrm{s} \overline{\mathbf{f}}, \\ 
\mathcal{F}_t &= \mathbf{A}_\textrm{g}\overline{\mathcal{F}} =\mathbf{A}_g\mathbf{A}_s\overline{\mathbf{f}} := \mathbf{A}\overline{\mathbf{f}}.
\end{align*}
Finally, our problem of obtaining load distribution can be simplified into a quadratic program (QP):
\begin{align}
    \text{min} ~& \overline{\mathbf{f}}^\top \mathbf{W} \overline{\mathbf{f}} \label{eqn:QP_forcedistribution}\\
    \textrm{subject to} ~& \mathcal{F}_t = \mathbf{A}\overline{\mathbf{f}}. \nonumber
\end{align}
It is known that a QP with only linear equality constraints can be solved analytically using Lagrange multipliers. By applying this method, we  compute the wrench distributed on each suction cup as follows:
\begin{equation} \label{eqn:load_dist_soln}
\overline{\mathcal{F}} = \mathbf{A}_s\overline{\mathbf{f}}=\mathbf{A}_s\mathbf{W}^{-1}\mathbf{A}^\top(\mathbf{A}\mathbf{}\mathbf{W}^{-1}\mathbf{A}^\top)^{-1}\mathcal{F}_t.
\end{equation} 

\subsection{Adjustable Stiffness Redistribution Model}
\label{subsec:redistribution}
When the suction cup is fully compressed (i.e., when the object being held ``bottoms out'' in the cup), the spring behavior enters a different phase, causing the stiffness to increase significantly as illustrated in Fig.~\ref{fig:pointforce_compressed}B. To account for this change, we resolve our optimization problem with higher stiffness coefficients. This redistribution ensures that the model accurately captures system behavior under varying compression levels in a computationally efficient manner.

To determine if the suction cup is likely to be fully compressed, we first identify the force distribution using standard weights. If the force meets the criterion for full compression, we adjust the weight corresponding to the force and resolve the problem. The detailed process is illustrated in the Algorithm \ref{alg:force_redistribution}. Additionally, the estimation of weights representing stiffness, both when the cup is fully compressed and when it is not, will be discussed in Section~\ref{sec:experiment}.


\begin{algorithm}[t]
\caption{Load distribution model with adjusting stiffness}\label{alg:alg1}
\begin{algorithmic}
\STATE \textbf{Input:} a wrench $\mathcal{F}_t$ exerted by a movement of an object \\
\STATE \textbf{Output:} a wrench distribution $\overline{\mathcal{F}}^*$ \\
\STATE initial force distribution
\STATE \hspace{0.5cm} set $\mathbf{W}$=diag$([\mathbf{W}_1, \cdots,  \mathbf{W}_{N_s}])$, where $\mathbf{W}_i=\mathbf{W}_\text{normal}$ 
\STATE \hspace{0.5cm} get $\mathcal{F}_i$ from $\overline{\mathcal{F}} = \mathbf{A}_s\mathbf{W}^{-1}\mathbf{A}^\top(\mathbf{A}\mathbf{W}^{-1}\mathbf{A}^\top)^{-1}\mathcal{F}_t$
\STATE identifying compressed suction cup and adjusting weights
\STATE \hspace{0.5cm} \textbf{for} $i=1\rightarrow {N_s}$ \textbf{do}
\STATE \hspace{1.0cm} \textbf{if} $\mathcal{F}_i[6] > f_{z,\text{threshold}} $ \textbf{then}
\STATE \hspace{1.5cm} $\mathbf{W}_i=\mathbf{W}_\text{compressed}$
\STATE \hspace{1.0cm} \textbf{end if}
\STATE \hspace{0.5cm} \textbf{end for}
\STATE redistribution
\STATE \hspace{0.5cm} set $\mathbf{W}$=diag$([\mathbf{W}_1, \cdots,  \mathbf{W}_{N_s}])$
\STATE \hspace{0.5cm} $\overline{\mathcal{F}}^* = \mathbf{A}_s\mathbf{W}^{-1}\mathbf{A}^\top(\mathbf{A}\mathbf{}\mathbf{W}^{-1}\mathbf{A}^\top)^{-1}\mathcal{F}_t$
\end{algorithmic}
\label{alg:force_redistribution}
\end{algorithm}

%
%

\section{Motion Planning}
\label{sec:mop}

\subsection{Planning Pipeline}
 Generating trajectories in real time is often challenging~\cite{chettibi2004minimum, diehl2006fast, schulman2013finding, zhao2018efficient, zhang2022time, wen2022path}. To achieve satisfactory results within reasonable computational limits, motion planning is often divided into two stages~\cite{shin1985minimum, shiller1989robot, shiller1991computing}.
In this paper, we also follow a two-step approach: 
\subsubsection{Path Planning} Generating a collision-free path considering kinematic constraints.
\subsubsection{Time-Optimal Trajectory Planning}
Creating a time-optimal trajectory that tracks the given path while satisfying kinodynamic constraints. 

We will skip the path planning part in this paper, assuming that the path is provided using state-of-the-art path-planning algorithms~\cite{kavraki1996probabilistic, lavalle1998rapidly, kuffner2000rrt, karaman2011sampling, devaurs2015optimal, gammell2015batch}. Instead, we focus on maintaining stable suction grasps during the trajectory generation stage. This problem can be defined as follows:

Given a path $\bq(s), s\in [0,1]$, find a monotonically increasing time scaling parameter $s(t) :[0,T] \rightarrow [0,1]$ that 
\begin{itemize}
    \item satisfies the initial state $(s_0,\dot{s}_0)=(0,0)$ and the final state $(s_\textrm{end},\dot{s}_\textrm{end})=(1,0)$, 
    \item minimizes the total travel time T along the path,
    \item respects third or lower-order kinematics constraints and dynamics constraints due to hardware limitations, and
    \item respects grasp failure constraints to ensure a secure grasp during the motion.
\end{itemize}

\subsection{Background: Time-Optimal Path Parameterization (TOPP) }
For the sake of solving the time-optimal trajectory generation problem that tracks the prescribed path, we consider the path $\bq(s)$ as a function of the scalar path coordinate $s\in[0,1]$. Then we can rewrite the first and second-order time derivatives of $\bq(s)$ as follows:
\begin{eqnarray}
\dot{\bq}(s) = \bq'(s)\dot{s}, \quad \ddot{\bq}(s)=\bq''(s)\dot{s}^2+\bq'(s)\ddot{s}.
\end{eqnarray}


Furthermore, it is known that a second-order time differential equation can be transformed into a first-order differential equation based on the relation introduced in \cite{pfeiffer1987concept},
\begin{eqnarray}
\ddot{s} = \frac{d\dot{s}}{dt} =  \frac{d\dot{s}}{ds}\frac{ds}{dt} = \dot{s}'\dot{s}=\frac{1}{2}(\dot{s}^2)' ,
\label{eqn:uandx}
\end{eqnarray}
which finally allows second-order equations to be formulated as the linear form: ($\dot{s}^2, \ddot{s}$) \cite{verscheure2009time}, \cite{pham2018new}.

\subsection{Equations of Motion}

\begin{figure}[t]
    \centering    \includegraphics[width=0.5\linewidth]{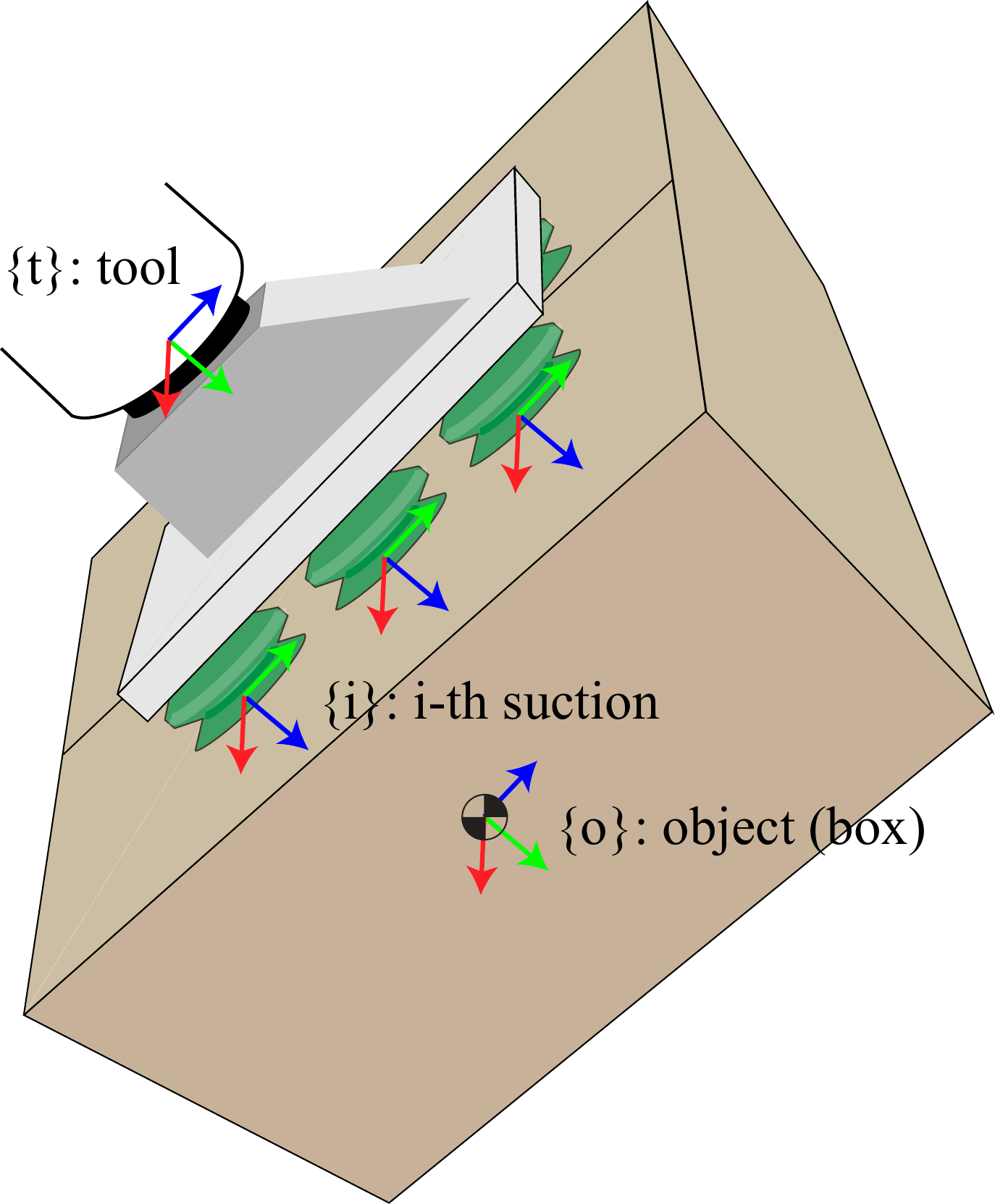}
    \caption{Frame description for our problem. Given that the object frame, attached to the CoM of the object, changes each time the robot handles different boxes, we will use the tool frame, attached to the end-effector of the manipulator, to formulate the problem.}
    \label{fig:suction}
    \vspace{-3mm}
\end{figure}

To define the grasp failure constraints, we first derive the equations of motions related to forces and moments applied to the suction cups by the movement of the payload object as shown in Fig \ref{fig:suction}.
Using Newton-Euler equation, we can state the translational and rotational dynamics of a rigid body via the sum of forces and moments acting on the rigid body. Then we obtain:
\begin{align}
\mathcal{F}_\textrm{o} &=  \sum_{i\in \text{suction}} [\textrm{Ad}_{\mathbf{T}_i^\textrm{o}}]^\top \mathcal{F}_i + \begin{bmatrix}0 \\ m_\textrm{o} \mathbf{g}_\textrm{o} \end{bmatrix} \nonumber \\
&= \begin{bmatrix} \mathbf{I}_\textrm{o} \balpha_\textrm{o} + \bomega_\textrm{o} \times \mathbf{I}_\textrm{o} \bomega_\textrm{o} \\ m_\textrm{o} \mathbf{a}_\textrm{o}\end{bmatrix}.
\label{eqn:ext_wrench_sum}
\end{align}
Note that we represent all the quantities as body wrenches and body twists, i.e., $\mathbf{v}_{o}$ represents the velocity of \{o\} expressed in \{o\} frame, not in the world frame. As the grasped box changes, the quantities related to the object are not fixed and must be estimated each time, whereas the quantities defined between the tool and each suction cup are fixed. Therefore, we reformulate the equation in the tool's frame. Without loss of generality, we align the orientation of the box frame with the tool's frame (i.e., $\mathbf{R}_t^\textrm{o} = \mathbf{I}_\mathrm{3\times3}$), allowing for a simpler description as:
\begin{align}
 \sum_{i\in \text{suction}} [\textrm{Ad}_{\mathbf{T}_i^\textrm{o}}]^\top \mathcal{F}_i &= [\textrm{Ad}_{\mathbf{T}_t^\textrm{o}}]^\top   \sum_{i\in \text{suction}} [\textrm{Ad}_{\mathbf{T}_i^t}]^\top \mathcal{F}_i \nonumber \\ 
    & = \begin{bmatrix} \mathbf{I}_\textrm{o} \balpha_t + \bomega_t \times \mathbf{I}_\textrm{o} \bomega_t \\ m_\textrm{o} ( \mathbf{a}_t - [\mathbf{p}_t^\textrm{o}]_\times\balpha_t )\end{bmatrix} - \begin{bmatrix} 0 \\ m_\textrm{o} \mathbf{g}_t \end{bmatrix}. 
     \nonumber
\end{align}

Finally, we get
\begin{align} 
     \mathcal{F}_t &= \sum_{i\in \text{suction}} [\textrm{Ad}_{\mathbf{T}_i^t}]^\top \mathcal{F}_i  \nonumber \\
     &=  [\textrm{Ad}_{\mathbf{T}_i^\textrm{o}}]^{-\top} \big( 
     \begin{bmatrix} \mathbf{I}_\textrm{o} \balpha_t + \bomega_t \times \mathbf{I}_\textrm{o} \bomega_t \\ m_\textrm{o} ( \mathbf{a}_t - [\mathbf{p}_t^\textrm{o}]_\times\balpha_t )\end{bmatrix} - \begin{bmatrix} 0 \\ m_\textrm{o} \mathbf{g}_t \end{bmatrix} 
     \big) \nonumber \\
     &= \footnotesize{ \begin{bmatrix} \mathbf{I}_\textrm{o} \balpha_t + \bomega_t \times \mathbf{I}_\textrm{o} \bomega_t 
    +m_\textrm{o} [\mathbf{p}_t^\textrm{o}]_\times( \mathbf{a}_t + [\mathbf{p}_t^\textrm{o}]_\times\balpha_t -\mathbf{g}_t)\\ 
    m_\textrm{o} ( \mathbf{a}_t + [\mathbf{p}_t^\textrm{o}]_\times\balpha_t -\mathbf{g}_t)\end{bmatrix}.  } \label{eqn:eom}
\end{align}

Now we aim to parameterize the equations of motion above as functions of $\dot{s}, \ddot{s}$. To achieve this, we first parameterize gravity, angular velocity, angular acceleration, and linear acceleration of the tool frame as follows:
\begin{align}
\mathbf{g}_t &= \mathbf{R}_t^w(\bq)\cdot[0, 0, -9.8]^\top, \nonumber \\
\bomega_t &= \mathbf{J}_t^{\bomega} (\bq(s))\bq'(s)\dot{s}, \nonumber \\
\balpha_t &= \mathbf{J}_t^{\bomega} (\bq) (\bq'\ddot{s}+\bq''\dot{s}^2)+ {\mathbf{J}}_t^{\bomega}(\bq,\bq')'\bq'\dot{s}^2, \nonumber  \\
\mathbf{a}_t &= \mathbf{J}_t^{\mathbf{v}} (\bq) (\bq'\ddot{s}+\bq''\dot{s}^2)+ \mathbf{J}_t^{\mathbf{v}}(\bq,\bq')'\bq'\dot{s}^2. 
\label{eqn:motion}
\end{align}

Then by substituting (\ref{eqn:motion}) into (\ref{eqn:eom}), forces experienced at each suction cup can be expressed as a function of the path parameter and its derivatives as follows:
\begin{equation}
\mathcal{F}_t = \mathbf{b}''(s) \ddot{s} + \mathbf{b}'(s) \dot{s}^2 + \mathbf{b}(s). \label{eqn:paramforce}
\end{equation}

\subsection{Grasp Failure as Second-Order Constraints in TOTP}
Our goal is now to represent grasping stability as a constraint in the TOTP problem, which may restrict the robot motion, i.e., limiting the magnitudes of $\dot{s}$ and $\ddot{s}$. As  derived in Section~\ref{sec:grasp_failure}, the grasp failure conditions can be represented as follows:
\begin{align}
 \overline{\mathbf{U}} \; \overline{\mathcal{F}} \leq \overline{\mathbf{u}} \tag{\ref{eqn:sucloss_ineq_full}}, \\
 \mathbf{U}_t \mathcal{F}_{t} \leq \mathbf{u}_t. \tag{\ref{eqn:slippage_ineq}} 
\end{align}
Then, by applying the obtained analytic solution $\overline{\mathcal{F}}$ from Equation (\ref{eqn:load_dist_soln}) and parameterized $\mathcal{F}_t$ from Equation (\ref{eqn:paramforce}) into (\ref{eqn:sucloss_ineq_full}) and (\ref{eqn:slippage_ineq}), we  obtain:
\small
\begin{align*}
    \overline{\mathbf{U}} \mathbf{A}_s\mathbf{W}^{-1}\mathbf{A}^\top(\mathbf{A}\mathbf{}\mathbf{W}^{-1}\mathbf{A}^\top)^{-1} \big( \mathbf{b}''(s) \ddot{s} + \mathbf{b}'(s) \dot{s}^2 + \mathbf{b}(s) \big) &\leq \overline{\mathbf{u}}  \\
    \mathbf{U}^t \big( \mathbf{b}''(s) \ddot{s} + \mathbf{b}'(s) \dot{s}^2 + \mathbf{b}(s) \big) &\leq \mathbf{u}^t,
\end{align*}
\normalsize
which can be combined and simplified as follows:
\begin{align}
   \bzeta''(s) \ddot{s} + \bzeta'(s) \dot{s}^2 + \bzeta(s) &\leq \mathbf{0}. \label{eqn:grasping_const_ineq}
\end{align}

\subsection{TOTP Constraints for the Discretized System}

We build upon the problem formulation described in our previous work \cite{lee2024performance}. We define the path parameter, divided into $N$ segments, as $0=:s_0, s_1,...,s_{N-1},s_N:=1$. For  convenience, we denote each quantity at $s_k$ as $(\cdot)(s_k)=(\cdot)_k$. We then set our optimization variables to $x_k=\dot{s}_k^2$ and rewrite the relation illustrated in Equation (\ref{eqn:uandx}) as follows:
\begin{equation*}
\ddot{s}_k = \frac{x_{k+1}-x_k}{2\triangle_k},\quad \textrm{where~}  \triangle_k :=s_{k+1}-s_k.
\end{equation*}

We then reformulate the grasping constraints ({\ref{eqn:grasping_const_ineq}) in the discretized system as follows:
\begin{align}
\bzeta''_k \ddot{s}_k + \bzeta'_k \dot{s}_k^2 + \bzeta_k  &\leq \mathbf{0} \nonumber \\
\Longleftrightarrow ~ 
  \frac{\bzeta''_k}{2\triangle_k}x_{k+1}  + (\bzeta'_k-\frac{\bzeta''_k}{2\triangle_k}) x_k + \bzeta_k  &\leq \mathbf{0}.
\end{align}

Additionally, we compute the force distribution given the nominal values of $x_{1:N}$ as:
\begin{align}
\overline{\mathcal{F}}_k &= \mathbf{A}_s\mathbf{W}^{-1}\mathbf{A}^\top(\mathbf{A}\mathbf{}\mathbf{W}^{-1}\mathbf{A}^\top)^{-1} (\mathbf{b}''_k \ddot{s}_k + \mathbf{b}'_k \dot{s}^2_k + \mathbf{b}_k) \nonumber \\
&:= \bphi_k^{''\mathbf{W}} x_{k+1}  + \bphi_k^{'\mathbf{W}} x_k + \bphi_k^{\mathbf{W}}.
\end{align}

Finally, all first- to third-order constraints, including the grasping stability constraints, can be formulated as a linear matrix inequality. The details can be found in our previous work\cite{lee2024performance}.
\small
\begin{numcases}{}
\textrm{1st order constraints:}&{$\bomega_k x_k \leq \bnu_k$} \nonumber\\
\textrm{2nd order constraints:}&{$\balpha_k^0 x_k + \balpha_k^1 x_{k+1} \leq \bbeta_k$ }\label{eqn:1and2and3} \\
\textrm{3rd order constraints:}&$\bgamma_k^0 x_k + \bgamma_k^1 x_{k+1} + \bgamma_k^2 x_{k+2} \leq \boeta_k$. \nonumber
\end{numcases}
\normalsize

\subsection{Trajectory Optimization}

In this subsection, we formulate the time-optimal trajectory planning problem based on the constraints derived above and describe the entire algorithm. The cost function for minimizing the total time required to follow the given path can be expressed as follows:
\begin{equation}
    f(\mathbf{x}) = \sum_{k=0}^{N-1} \frac{\triangle_{i}}{\sqrt{x_k}+\sqrt{x_{k+1}}},
\end{equation}
which can be linearized along the nominal trajectory $\overline{\mathbf{x}}$. 

We now formulate the trajectory optimization problem as a Sequential Linear Program (SLP), where the optimization variables are $\mathbf{x} = [x_1, \cdots, x_{N-1}]^\top$ as follows: 
\begin{align}
& \min_{\mathbf{x}} \quad \mathbf{c}^\top\mathbf{x}  \label{eqn:opt} \\
& \textrm{subject to}\quad \mathbf{A} \mathbf{x} \leq \mathbf{b}, \nonumber
\end{align}
where
\begin{equation}\label{eqn:cAb}
\mathbf{c} = \frac{\partial f}{\partial \mathbf{x}}\bigg\vert_{\mathbf{x}=\mathbf{\overline{x}}}, \; \mathbf{A} = \begin{bmatrix}  A^1 \\ A^2 \\ A^3  \end{bmatrix},\; \mathbf{b} = \begin{bmatrix}  b^1 \\ b^2 \\ b^3 \end{bmatrix},
\end{equation} 
with $\mathbf{A}^i, \mathbf{b}^i$ representing inequality coefficients for the stacked $i$th-order constraints formulated as:

{\footnotesize
\begin{align*}
A^1 &= \begin{bmatrix}
\bomega_1 & 0 & \cdots & 0 \\
\vdots & &  & \vdots \\
 0 & \cdots & 0 & \bomega_{N-1} 
\end{bmatrix}, %
\quad\;\; b^1 = \begin{bmatrix}
\bnu_1 \\ \vdots \\ \bnu_{N-1} 
\end{bmatrix}, \nonumber \\
A^2 &= \begin{bmatrix}
\balpha^1_{0} & 0 & \cdots & 0 \\
\balpha^0_{1} & \balpha^1_{1} & \cdots & 0 \\
\vdots & & & \vdots \\
0 & \cdots & \balpha^0_{N-2} & \balpha^1_{N-2}   \\
0 & \cdots & \cdots & \balpha^0_{N-1}   
\end{bmatrix}, %
b^2 = \begin{bmatrix}
\bbeta_{0}-\balpha_0^0x_0 \\
\bbeta_{1} \\ \vdots \\ \bbeta_{N-2} \\ 
\bbeta_{N-1} - \balpha_{N-1}^1x_N
\end{bmatrix}, \nonumber \\
A^3 &= \begin{bmatrix}
\bgamma^2_0 & 0 & 0 & 0 & \cdots & 0 \\
\bgamma^1_1 & \bgamma^2_1 & 0 & 0 & \cdots & 0 \\
\bgamma^0_2 & \bgamma^1_2 & \bgamma^2_2 & 0 & \cdots & 0 \\
\vdots & & \vdots & \ddots & \ddots & \vdots \\
0 & \cdots & 0 & \bgamma^0_{N-3} & \bgamma^1_{N-3} & \bgamma^2_{N-3}\\
0 & \cdots & 0 & 0 & \bgamma^0_{N-2} & \bgamma^1_{N-2}
\end{bmatrix},  \\ 
b^3 &= \begin{bmatrix}  \boeta_0 - \bgamma^0_0 x_0 -\bgamma^1_0 x_1 \\
 \boeta_{1} - \bgamma^0_{1}x_1 \\
 \boeta_{2} \\ \vdots \\ \boeta_{N-3} \\ \boeta_{N-2} - \bgamma^2_{N-2} x_N
\end{bmatrix}.
\end{align*}
}

Note that due to linearizing, we need to iteratively update the nominal trajectory to reformulate the coefficients for the cost function and 3rd-order constraints as described in \cite{lee2024performance}. Additionally, as outlined in Algorithm~\ref{alg:force_redistribution} from Section~\ref{subsec:redistribution}, we aim to adjust the stiffness coefficients for the suction cup likely to experience significant compression. This adjustment is based on the initial force distribution prediction computed using normal weights. These updates are applied during each iteration, as shown in Algorithm~\ref{alg:TOTP_grasp}.

\begin{algorithm}[t!]
\caption{TOTP3 with grasp constraints}\label{alg:TOTP_grasp}
\begin{algorithmic}
\STATE \hspace{-0.3cm} {\textbf{Require}} $\overline{\mathbf{x}}_{1:N}$
\STATE \hspace{0cm}\textbf{While} True \textbf{do}
\STATE \hspace{0.3cm}\textbf{For} $k=1\longrightarrow N$ \textbf{do}
\STATE \hspace{0.6cm}initial force distribution:
\STATE \hspace{0.9cm}set $\mathbf{W}=$\;diag$([\mathbf{W}_1,..,\mathbf{W}_{N_s}])$, where $\mathbf{W}_i=\mathbf{W}_\text{normal}$ 
\STATE \hspace{0.9cm}compute $\overline{\mathcal{F}} =\bphi_k^{''\mathbf{W}} \bar{x}_{k+1}  + \bphi_k^{'\mathbf{W}} \bar{x}_k + \bphi_k^{\mathbf{W}}$
\STATE \hspace{0.6cm}identifying compressed cup and adjusting weights: 
\STATE \hspace{0.9cm}\textbf{For} $i=1\rightarrow {N_s}$ \textbf{do}
\STATE \hspace{1.2cm}get $\mathcal{F}_i$ from $\overline{\mathcal{F}}$ 
\STATE \hspace{1.2cm}$\mathbf{W}_i=\mathbf{W}_\text{compressed}$ \textbf{if} $\mathcal{F}_i[6] > f_{z,\text{threshold}}$
\STATE \hspace{0.9cm}\textbf{End For}
\STATE \hspace{0.9cm}set $\mathbf{W}$=diag$([\mathbf{W}_1, \cdots,  \mathbf{W}_{N_s}])$
\STATE \hspace{0.6cm}update grasping constraints: $\bzeta''_k, \bzeta_k, \bzeta_k$
\STATE \hspace{0.6cm}update 2nd-order constraints: $\balpha^0_k, \balpha^1_k, \bbeta_k$
\STATE \hspace{0.6cm}update 3rd-order constraints: $\bgamma^0_k, \bgamma^1_k,\bgamma^2_k, \boeta_k$
\STATE \hspace{0.3cm}\textbf{End For}
\STATE \hspace{0cm}update $\mathbf{c}, \mathbf{A}, \mathbf{b}$ \hfill{Equation \eqref{eqn:cAb}}
\STATE \hspace{0cm}$\mathbf{x}_{1:N}$ = solve LP \hfill{Equation \eqref{eqn:opt}}
\STATE \hspace{0cm}\textbf{If} {$\|\mathbf{x}_{1:N}-\overline{\mathbf{x}}_{1:N}\| < \epsilon$} break
\STATE \hspace{0cm}\textbf{End If}
\STATE \hspace{0cm}Update $\overline{\mathbf{x}}_{1:N}\longleftarrow\mathbf{x}_{1:N}$
\STATE \hspace{-0.3cm} \textbf{End While}
\end{algorithmic}
\end{algorithm}


%
%

\section{Experiment Results}
\label{sec:experiment}

\subsection{Gripper Testbed}
To evaluate our assumptions regarding the grasp failure constraints, we developed a testbed capable of measuring the total gripper wrench as well as the wrenches exerted at each suction cup, as depicted in Fig.~\ref{fig:exp_setup}. The testbed included six suction cups and was bolted to the floor to allow controlled application of forces and wrenches to the gripper and its cups. To measure the force and torque applied to each suction cup, we used ``ATI Axia80-M50" force/torque (F/T) sensors attached to the bottom of each suction cup and an ``ATI Axia130-M125" F/T sensor at the base to measure the total wrench applied to the overall test gripper.

For each suction cup, we used the Schmalz ``SPB1 60 ED-65 G1/4-IG" suction cup along with ``SCPSi-L HV 3-20 NC M12-5" venturi-based vacuum generator. Per manufacturer specifications, each suction cup features 1.5 bellows, a diameter of \mbox{60 mm}, and can produce a suction force of \mbox{78 N} as well as a pull-off force of \mbox{100.9 N} at a vacuum level of \mbox{600 mbar}. Furthermore, we confirmed that the vacuum level is in the range of \mbox{912-922 mbar} for our operating pressure of \mbox{6 bar} based on the venturi specifications. From this, we estimated that our 6-cup gripper testbed would generate an approximate suction force of \mbox{118.6 N} and a pull-off force of \mbox{155 N} for each suction cup. Usually, the pull-off force is higher than the suction force due to the adhesive forces between the suction cup and the surface, as well as the deformation of the suction cup creating slight additional resistance.

\begin{figure}[t]
    \centering
    \vspace{-2mm}
    \includegraphics[width=0.95\linewidth]{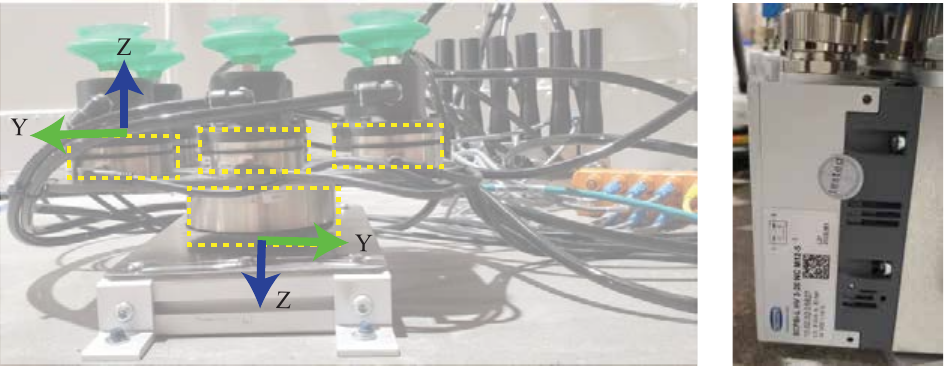}
    \vspace{-3mm}
    \caption{A testbed gripper and the vacuum generator used in our experimental setup. F/T sensors were attached to the bottom of each suction cup and to the base to measure the total wrench applied to the tool. }
    \label{fig:exp_setup}
    \vspace{-2mm}
\end{figure}

\begin{figure}[t]
    \centering
    \includegraphics[width=0.95\linewidth]{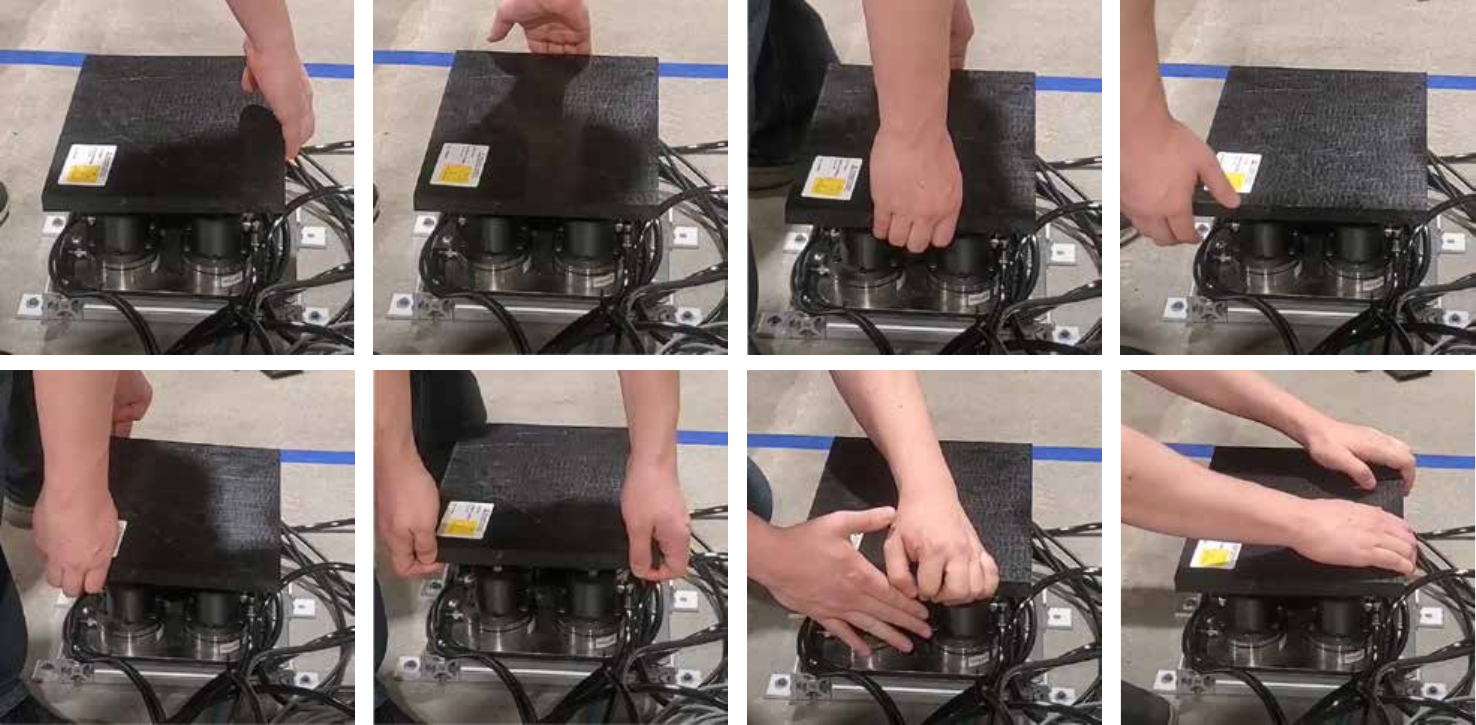}
    \vspace{-2mm}
    \caption{Snapshots of load distribution measurements showing the varying forces generated across different points on the surface in multiple directions.}
    \label{fig:snapshots_frcdist}
    \vspace{-5mm}
\end{figure}

\subsection{Load Distribution}
To assess the accuracy of our load distribution model, we applied wrenches to the grasping object in different ways by altering the magnitude, direction, and point of application of forces as illustrated in Fig.~\ref{fig:snapshots_frcdist}. During the application of these wrenches, we collected synchronized wrench data from the gripper base and suction-cup F/T sensors. By identifying the total wrench measured from the base as the load exerted by a box in our model, we could compute the load distribution based on the proposed Algorithm~\ref{alg:force_redistribution}
 derived in the previous section. We could then verify the accuracy of our model by comparing the predicted load distribution and the measured wrenches on each suction cup. 

\begin{figure*}[htb!]
\centering   
\includegraphics[width=0.95\linewidth]{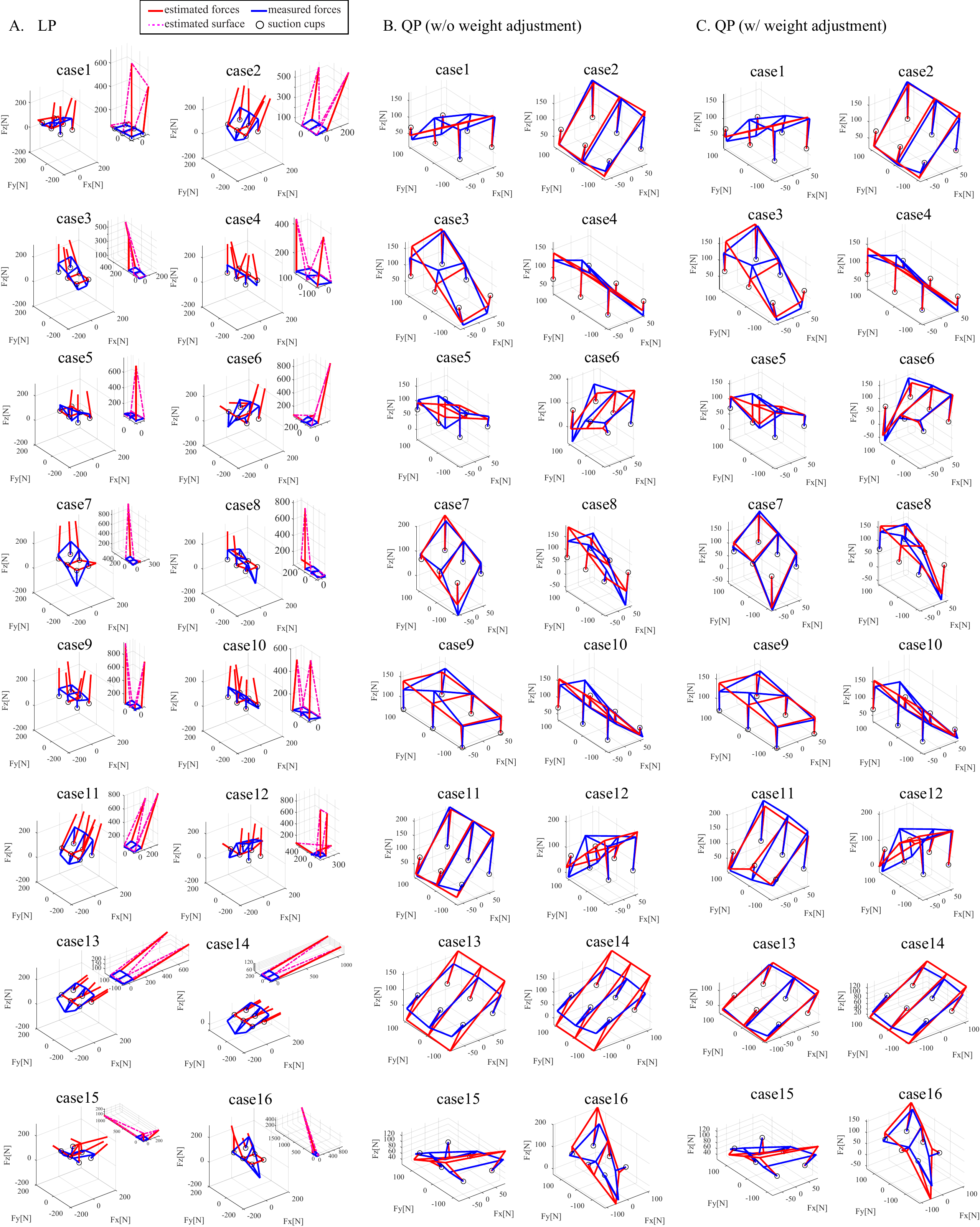}
 \caption{Visualized comparison of measured and estimated force distribution obtained by three different algorithms: A. LP (Linear Programming) load distribution, which solves for minimum stretched distance. Each figure includes a full graph on the right, which captures full forces exceeding the axis limits as well. B. QP (Quadratic Programming without weight adjustment) load distribution, which solves for minimum potential energy. C. QP (Quadratic Programming with weight adjustment) load distribution, which solves for minimum potential energy considering the compression of cups. It shows Linear Programming (LP) tends to distribute force being concentrated at specific points on the surface. While Quadratic Programming (QP) spreads force more evenly across the surface, resulting in a more uniform distribution. }
\label{fig:force_distribution_sufplot}
\end{figure*}

\subsubsection{Weight parameter estimation}
As shown in Algorithm~\ref{alg:force_redistribution}, our load distribution model requires estimates of the weight parameters $\mathbf{W}_\text{normal}$ and $\mathbf{W}_\text{compressed}$. Since our suction cups were axially symmetric, we assumed the stiffness was the same in the horizontal direction (parallel to the grasped surface) but varied in the normal direction (perpendicular to the grasped surface). We estimated the weights to fit the model to the data as closely as possible using a genetic algorithm~\cite{golberg1989genetic, conn1991globally, conn1997globally} implemented in MATLAB~\cite{GlobalOptimizationToolbox}. By fixing $w_\textrm{normal,xy}=1$ and setting the optimization variables $\mathbf{x}=[w_\textrm{normal,z}, w_\textrm{compressed,xy}, w_\textrm{compressed,z}, f_\text{z,threshold}]$, we could find the weight parameters that minimize the sum of force errors through the genetic algorithm. The optimal set of weights were found to be $\mathbf{W}_\text{normal}=\text{diag}([1.0, 1.0, 2.3682]), \mathbf{W}_\text{compressed}=\text{diag}([0.8369, 0.8369, 0.1321])$ with \mbox{$f_\text{z,threshold} = -47.19$ N} for our experimental setup. The resulting force distributions with and without weight adjustment, which we describe in Algorithm~\ref{alg:alg1} are shown in Fig. \ref{fig:force_distribution_sufplot} 
  and Fig. \ref{fig:force_distribution_plot}.


\begin{figure}[t]
\centering   
\includegraphics[width=0.95\linewidth]{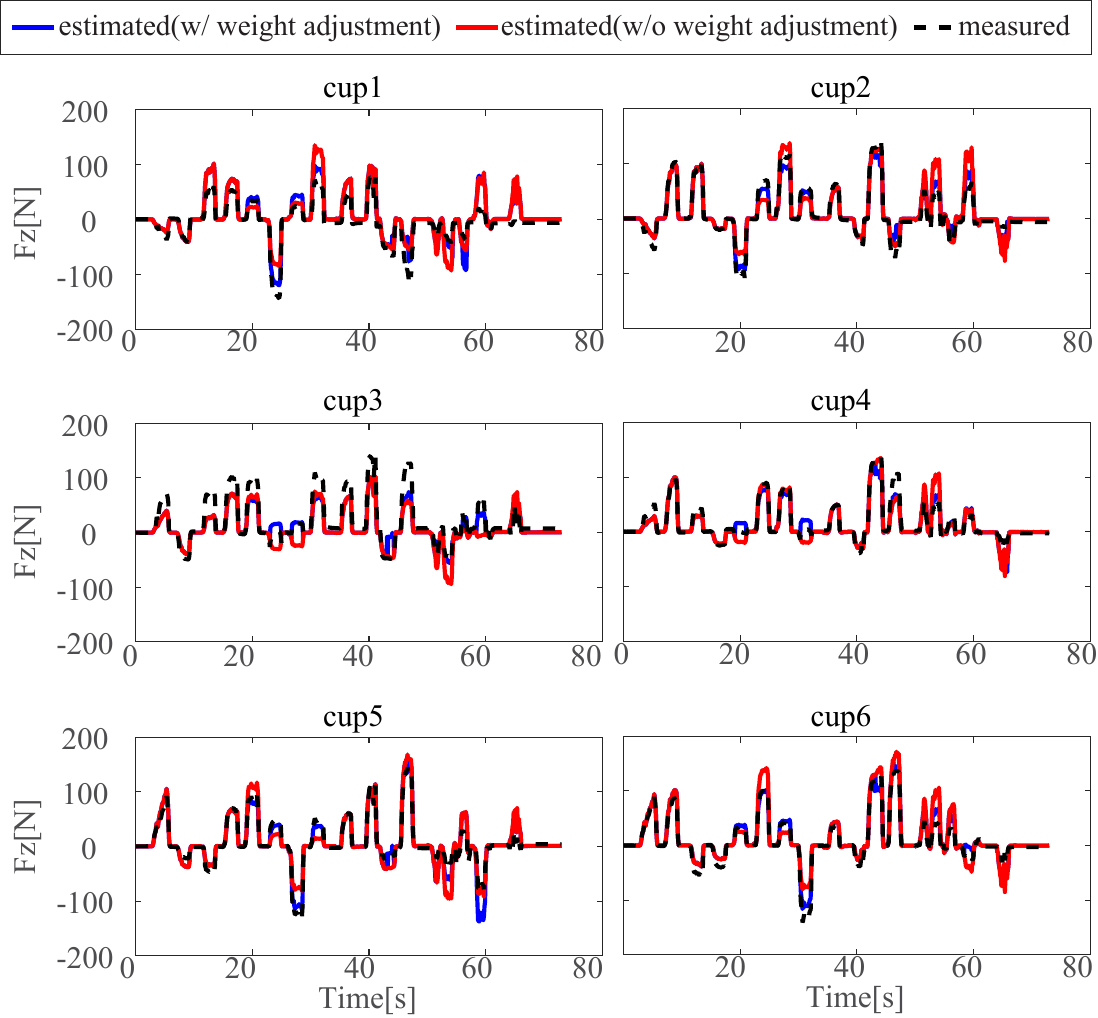}
\vspace{-2mm}
 \caption{Measured and estimated normal forces ($F_z$) applied to the six cups during the experiment. The results show that the estimated load distribution with weight adjustment (blue) has closer alignment with the measured real-world data (black dashed) than the one without weight adjustment (red).}
\label{fig:force_distribution_plot}
\vspace{-3mm}
\end{figure}

\subsubsection{Visualization}
To provide intuitive visualizations of the applied wrenches and the resulting force distribution on each suction cup, Fig.~\ref{fig:force_distribution_sufplot} presents plots of 3D forces exerted at each suction cup for different cases. In these plots, force vectors of neighboring cups are connected with lines to form surfaces, facilitating the visualization of force distribution. Each case is generated by capturing the instance when maximum wrenches were applied, as shown by the peaks in Fig.~\ref{fig:force_distribution_plot}, with some examples illustrated in Fig.~\ref{fig:snapshots_frcdist}. 

\subsubsection{Analysis of results}
In Fig.~\ref{fig:force_distribution_plot} we depict the time series of the suction cup forces recorded throughout the entire experiment. In particular, we focus on the normal direction of the forces ($F_z$), which holds the most significance regarding our study. The plot confirms that our model generally aligns well with real-world observations. However, it is worth noting that the data for cup 3 appears to deviate a bit more compared to other cups, which may be attributed to factors such as variations in suction cup height, which we measured and found to stem from manufacturing errors.

Fig.~\ref{fig:force_distribution_sufplot}A displays the distribution of estimated forces obtained through a linear programming (LP) algorithm, included for comparison against our QP-based approach. This formulation can be understood as the result of minimizing the total compression or tension exerted at each suction cup. The detailed formulation used for the analysis is described in the Appendix. The plots reveal that the LP approach leads to a concentration of forces in a single area, which diverges significantly from real-world observations. 

Next, Fig. \ref{fig:force_distribution_sufplot}B depicts the distribution of estimated forces that minimizes the spring energy of suction cups, as described in Equation~(\ref{eqn:QP_forcedistribution}). Considering the complex physics of the suction cups and the potential for insufficient accuracy in the measured data, the estimated force distribution appears quite similar to our measurements. However, we still observe slight unmatched discrepancies in cases 1, 3, 5-10, 12, and 15-16. 

Fig.~\ref{fig:force_distribution_sufplot}C depicts the distribution with weight adjustment as proposed in Algorithm~\ref{alg:force_redistribution}. By adjusting the weight of suction cups, which were considered to be in compression and thus expected to have different stiffness, we demonstrate an improved prediction of the force distribution. Although there are still some discrepancies (cases 1, 12, and 15), this weight adjustment method provides better results in most cases.

\begin{figure}[t]
    \centering
\includegraphics[width=0.95\linewidth]{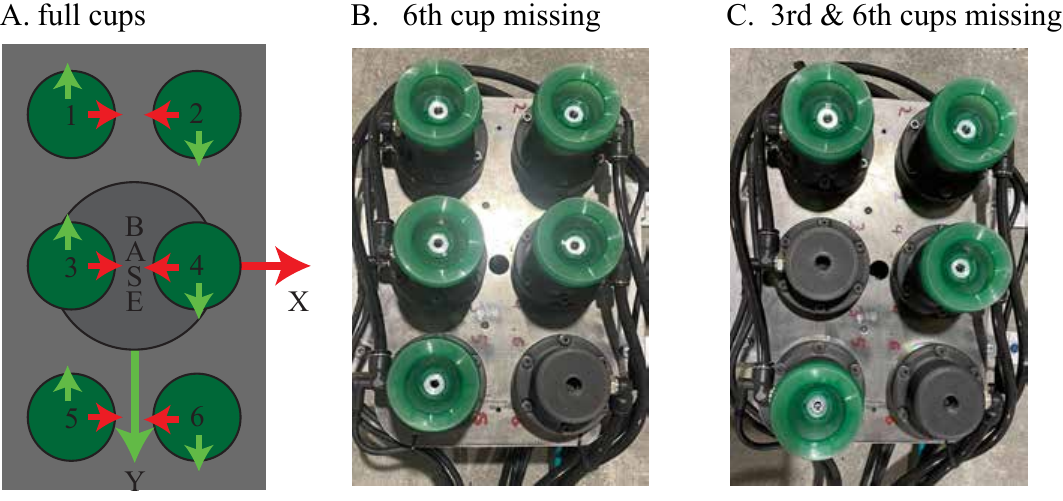}
\vspace{-3mm}
    \caption{Different gripper suction-cup arrangements for generalization validation:  A. A gripper with all the suction cups attached, B. A gripper missing the 6th suction cup, and C. A gripper missing two suction cups (3rd and 6th) to test grasp scenarios with various asymmetric arrangements.}
    \label{fig:diff_gripper}
    \vspace{-3mm}
\end{figure}

\begin{figure}[t]
    \centering
\includegraphics[width=0.95\linewidth]{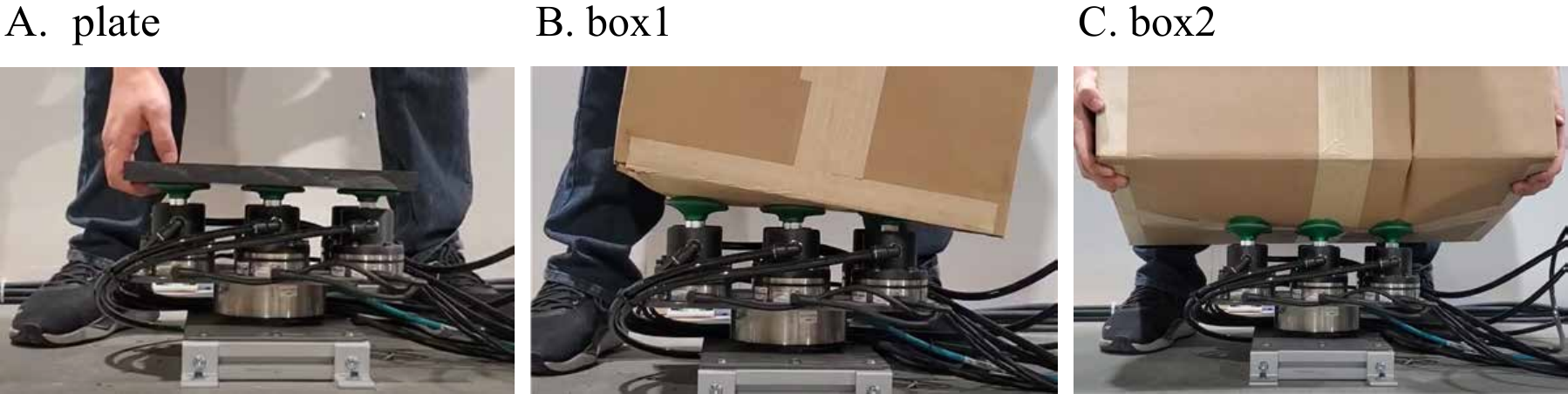}
\vspace{-2mm}
    \caption{The figure displays the different objects used in the experiment to assess the model's generalization ability. These objects vary in shape, size, and material to provide a diverse evaluation of the model's performance. }
    \label{fig:diff_objects}
    \vspace{-5mm}
\end{figure}

\begin{figure*}[!htb]
    \centering
\includegraphics[width=0.9\linewidth]{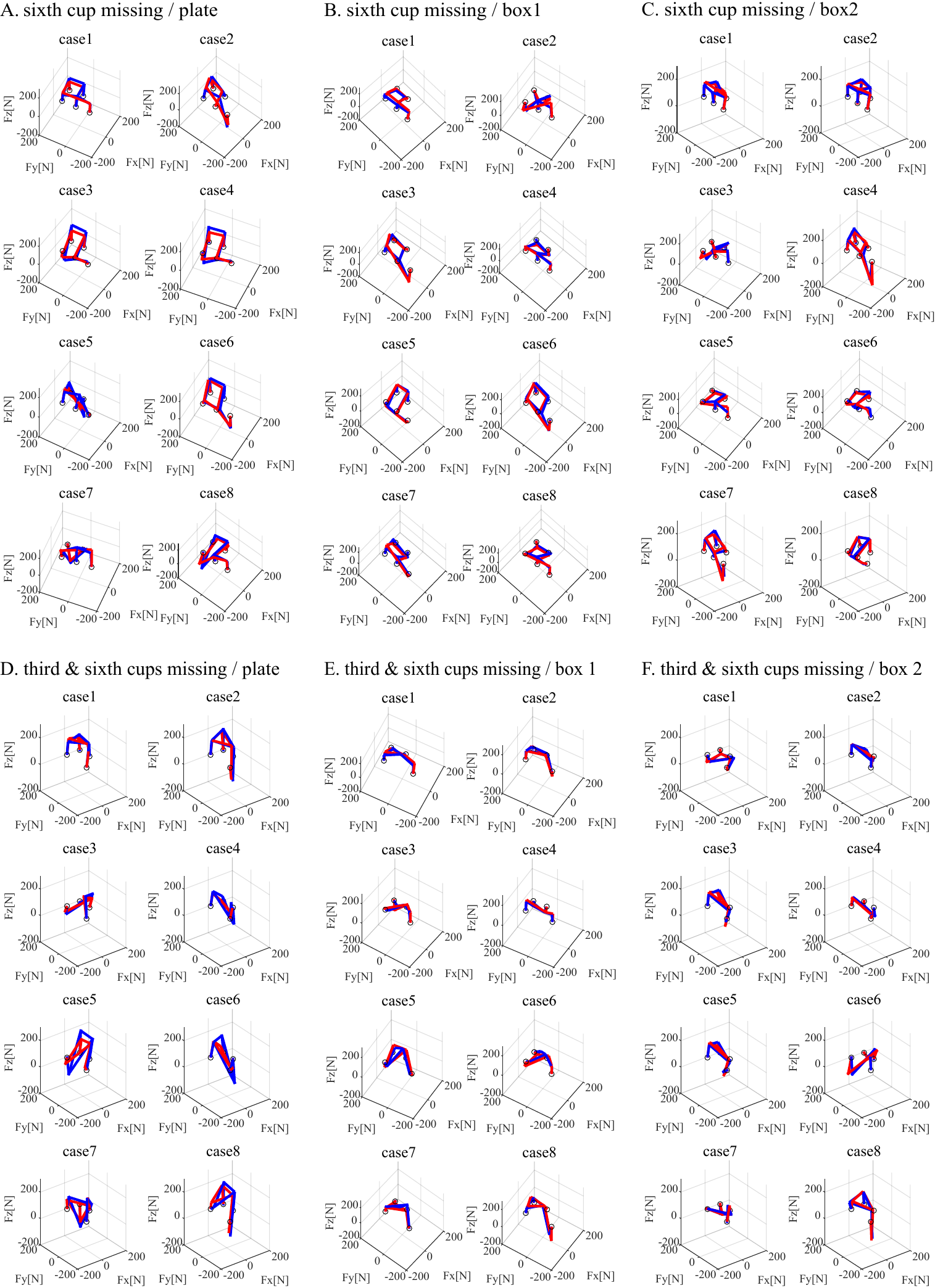}
\vspace{-2mm}
    \caption{Visualized comparison of measured and estimated force distribution for grippers with the sixth cup missing (See Fig.~\ref{fig:diff_gripper}B) and the third and sixth cups missing (See Fig.~\ref{fig:diff_gripper}C) handling various objects (See Fig.~\ref{fig:diff_objects}): The figure shows that the estimated load distribution (red lines) obtained by the proposed method can predict the actual load distribution measured by F/T sensor (blue lines) well.  }
    \label{fig:var_grip_obj}
\end{figure*}

\subsection{Generality of the proposed force distribution model}
One of the key strengths of our force distribution model is its generality due to its grounding in physical principles. By leveraging these principles, the model can predict force distribution and maintain grasp for a wide range of scenarios. This robustness and adaptability highlights the model's effectiveness in real-world applications, where the physical properties of the objects being manipulated are diverse and unpredictable.

\subsubsection{Different gripper configurations}
We also validated that our model could be used for grippers having a variety of suction cup arrangements. As shown in Fig~\ref{fig:diff_gripper}, we tested our algorithm with several gripper configurations including (1) having all suction cups attached, (2) missing one suction cup, and (3) missing two suction cups. The removed cups were selected to test the model's adaptability to asymmetric arrangements.

\begin{figure*}[t]
    \centering
    \vspace{-1mm}
    \includegraphics[width=0.95\linewidth]{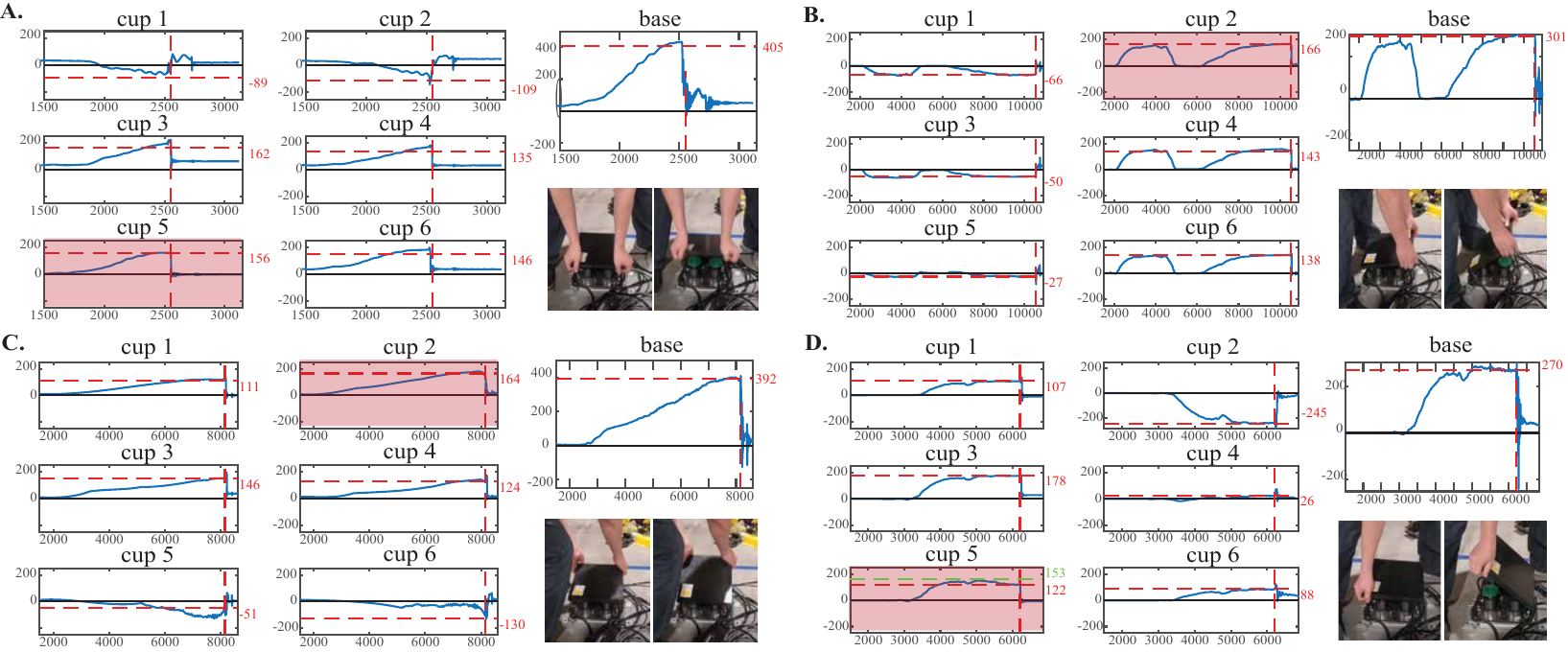}
    \vspace{-2mm}
    \caption{ \textbf{Suction loss failure test}: Plots of $F_z$ at each F/T sensor measured during the suction loss failure tests. In cases A through D, we applied and measured forces in different ways. Timing of suction loss is illustrated in red dotted lines and the corresponding force at that moment is represented in red on the right side of the plot. In each case, we mark the area in red corresponding to the cup where the suction grasp failure occurs. Additionally, snapshots for each case have been included to illustrate how we applied the forces.}
    \label{fig:plot_suction_loss}
    \vspace{-5mm}
\end{figure*}

\subsubsection{Handling diverse objects} Building on the experiments with diverse gripper configurations, we conducted a second series of tests to evaluate the model's performance when grasping diverse types of objects. Given that many tasks in warehouse automation involve box handling, we specifically evaluated the model's performance on two different sizes of boxes, as shown in Fig~\ref{fig:diff_objects}. 


\subsubsection{Analysis of results} In Fig.~\ref{fig:var_grip_obj}, we compare the estimated force distributions (red) obtained by the proposed method with the F/T sensor data (blue) to verify how well the model matches the real data. Initially, we used a gripper with the 6th cup missing, as shown in Fig. \ref{fig:diff_gripper}B, to handle a variety of objects (A: plate, B: box1, C: box2, See Fig. \ref{fig:diff_objects}). We then applied a series of forces via the grasped objects, simulating the forces generated during their movement when they are manipulated by a robot.
After completing experiments with the gripper missing the 6th cup, we conducted a similar experiment with a gripper missing the 3rd and 6th cups, as depicted in Fig. \ref{fig:diff_gripper}C. 

Finally, we compare the applied wrench distribution over each suction cup with the predictions from our model, as shown in Fig.~\ref{fig:var_grip_obj}. The plots demonstrate that our model accurately predicts the force distribution for two different grippers in asymmetrical configurations. This result reinforces the effectiveness of our method in estimating load distribution, a critical factor for employing grasp failure constraints in multi-suction-cup grippers with various suction-cup configurations. This approach can enhance the model's robustness and general applicability to real-world robotic gripping tasks, confirming its effectiveness across a wide range of gripper designs.

\subsection{Validation of the Grasp Failure Model}
\subsubsection{Suction Loss Constraints}

To identify the condition when a suction gripper loses its suction from an object being manipulated, we measured the forces exerted at each suction cup and the total force applied at the gripper's base while attempting to detach the plate from the gripper. Fig.~\ref{fig:plot_suction_loss}. depicts the plots of $F_z$ recorded at each F/T sensor during the experiment. From this data, we identified the pulling force $F_z$ which resulted in suction loss. Our observations indicated that the testbed gripper was prone to losing its suction primarily when the pulling force applied to any individual suction cup, especially those at the edge, exceeded their pull-off force limits (approximately \mbox{155 N}). This was more significant than the combined suction forces of all suction cups attached to the gripper (approximately \mbox{720-930 N}). 
This occurred because when one suction cup lost its suction, the force applied to that cup was redistributed to the remaining cups. This sudden increase in pulling force at the other cups can be observed in Fig.~\ref{fig:plot_suction_loss} and led to grasp failure from the remaining cups. In case D, the force at which suction failure occurred was measured at \mbox{122 N}, significantly lower than the anticipated value of around \mbox{155 N}. However, we observed that the maximum force applied to cup 5 before grasp failure was \mbox{153 N}. This suggested that the force causing suction failure likely exceeded \mbox{153 N}, but the F/T sensor may not have detected this due to the smoothing effect of the filter integrated into the sensor.

\subsubsection{Slippage Constraints}
\begin{figure}[t]
    \centering    \includegraphics[width=0.98\linewidth]{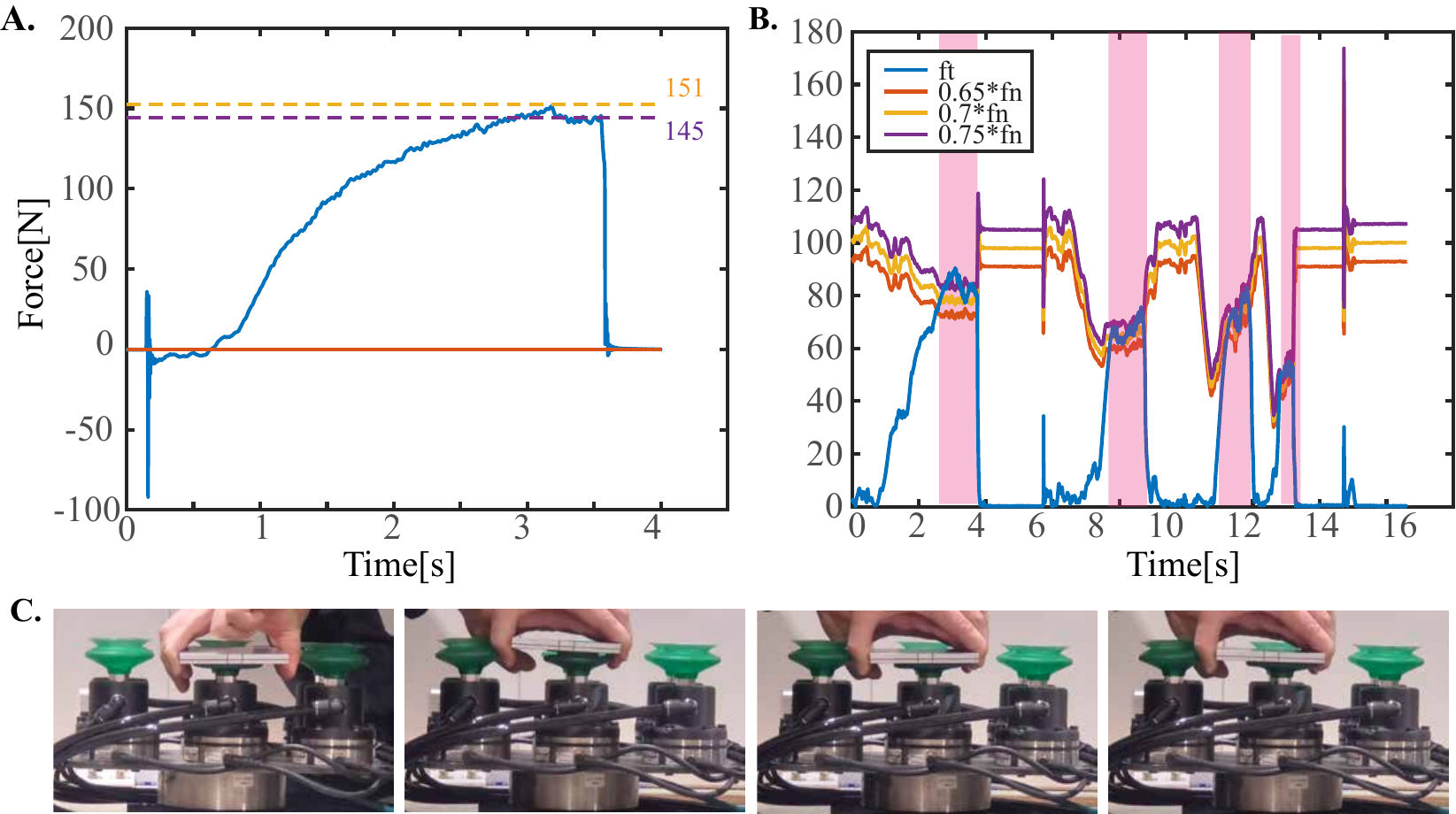}
    \vspace{-2mm}
    \caption{ \textbf{Identification of the Friction Coefficient for a Single Suction Cup}: Figure A identifies the pull-off force of a suction cup, and Figure B compares the tangential force measured from an F/T sensor with the frictional resistance (FR), calculated as the product of the friction coefficient and the normal force. The friction coefficient is then estimated by finding the point where the FR matches the tangential force, indicating the onset of slippage. }
    \label{fig:plot_singlecup}
    \vspace{-5mm}
\end{figure}

To identify the variables necessary to determine a slippage failure including suction force and a friction coefficient, we conducted a simple test using a single cup, as shown in Fig.~\ref{fig:plot_singlecup}. First, to measure the suction force, we applied gradually increasing pulling forces while recording the force in the $z$-direction, as shown in Fig.~\ref{fig:plot_singlecup}\textbf{A}. The graph shows that the suction force nearly reached equilibrium before pull-off, measuring approximately \mbox{145 N} for the cup. Next, to determine the static friction coefficient at which slippage begins, we applied force in a tangential direction until slippage occurred, as shown in snapshots of Fig.~\ref{fig:plot_singlecup}\textbf{C}. As depicted in Fig.~\ref{fig:plot_singlecup}\textbf{B}, we assumed a friction coefficient of $\mu=0.7$, which aligned with the observed data. 
 
\begin{figure}[t]
    \centering    \includegraphics[width=\linewidth]{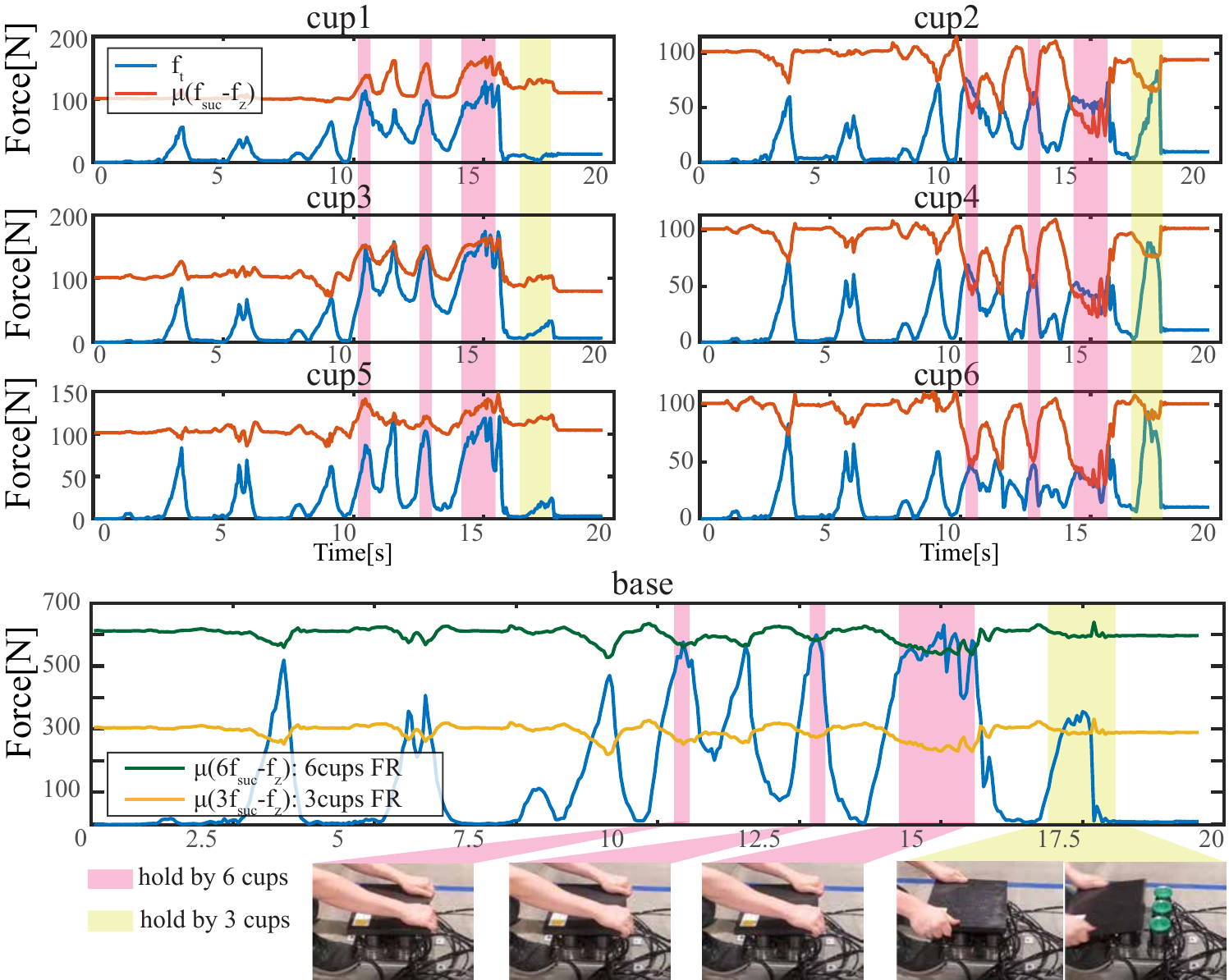}
    \vspace{-5mm}
    \caption{ \textbf{Slippage failure test}: Plots of applied tangential force $f_t$, and friction resistance (FR) $\mu(f_\textrm{suc}-f_z)$ at each F/T sensor measured during slippage failure tests. The corresponding snapshot shows that slip occurs when the total tangential force (blue line) applied to the object exceeds the friction resistance (green or yellow line) computed for the activated suction cups. }    \label{fig:plot_slippage}
    \vspace{-5mm}
\end{figure}

For slippage, we hypothesized a less conservative scenario compared to suction loss. We assumed that slippage is more likely when the total tangential force exceeds the friction resistance rather than focusing on forces applied to each suction cup. To verify this hypothesis, we performed a slippage failure test. As illustrated in Fig.~\ref{fig:plot_slippage}, we applied a tangential force to the plate to induce slip from the suction cup. During the experiment, we computed the tangential force as the sum of the $x$- and $y$-directional forces measured at the F/T sensors. Given the friction coefficient $\mu=0.7$ and a pre-measured suction force of \mbox{145 N}, we calculated the friction resistance using $\mu(f_\textrm{suc}-f_z)$, where $(f_\textrm{suc}-f_z)$ represents the normal force. 

As shown in Fig.~\ref{fig:plot_slippage}, we observed two short slippages at around \mbox{11 s} and \mbox{13 s} followed by two more significant slippages at around \mbox{14 s} and \mbox{17 s}, which aligned with the recorded data. During the initial major slippage around \mbox{15 s}, cups 1, 3, and 5  lost suction. The plots indicate significant slippage resistance at cups 1, 3, and 5, where the slippage condition was not met, ultimately resulting in suction loss in these cups. As a result, the final slippage occurred when the tangential force exceeded a threshold calculated assuming suction forces only from these three cups, as highlighted in yellow on the plot.

\subsection{Robot Experiments}

\subsubsection{Robot setup}

For the robot experiment, we used a 7-DOF robotic system composed of a 6-DOF RS020N Kawasaki arm mounted on an external revolute joint, as illustrated in Fig.~\ref{fig:robot_setup}A. A vacuum gripper equipped with 8 suction cups, configured as shown in Fig.~\ref{fig:robot_setup}B, was used to handle various sizes and weights of the boxes (Fig.~\ref{fig:robot_setup}C).

\subsubsection{Experiment setup}

\begin{figure}[t]
    \centering    \includegraphics[width=0.98\linewidth]{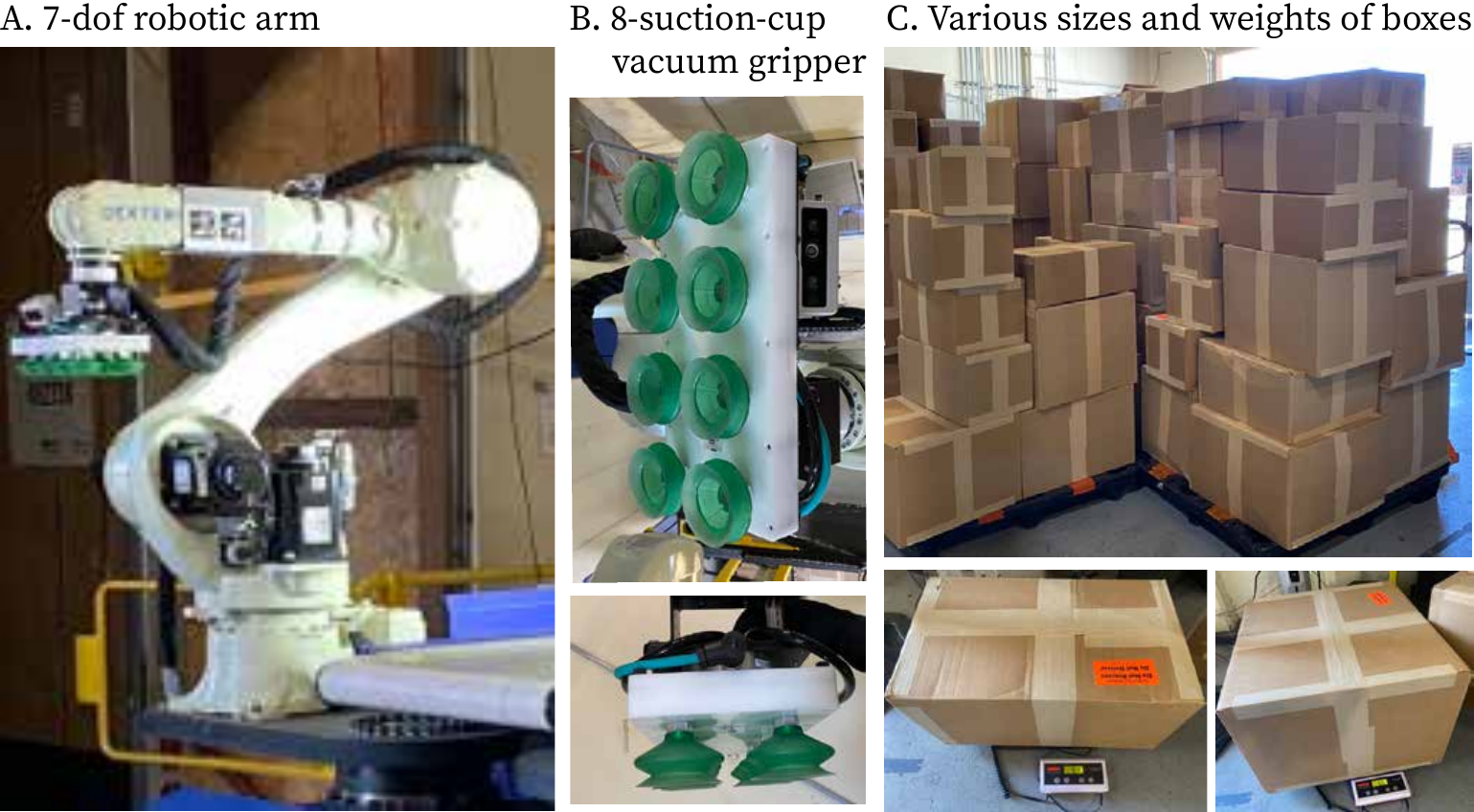}
    \vspace{-3mm}
    \caption{Robot experiment setup. A:  7-DOF robotic system composed of a 6-DOF RS020N Kawasaki arm mounted on an external revolute joint. B: A vacuum gripper equipped with 8 suction cups. C: Different sizes and weights of boxes used for the experiment.}
    \label{fig:robot_setup}
    \vspace{-3mm}
\end{figure}
\begin{figure}[t]
    \centering    \includegraphics[width=\linewidth]{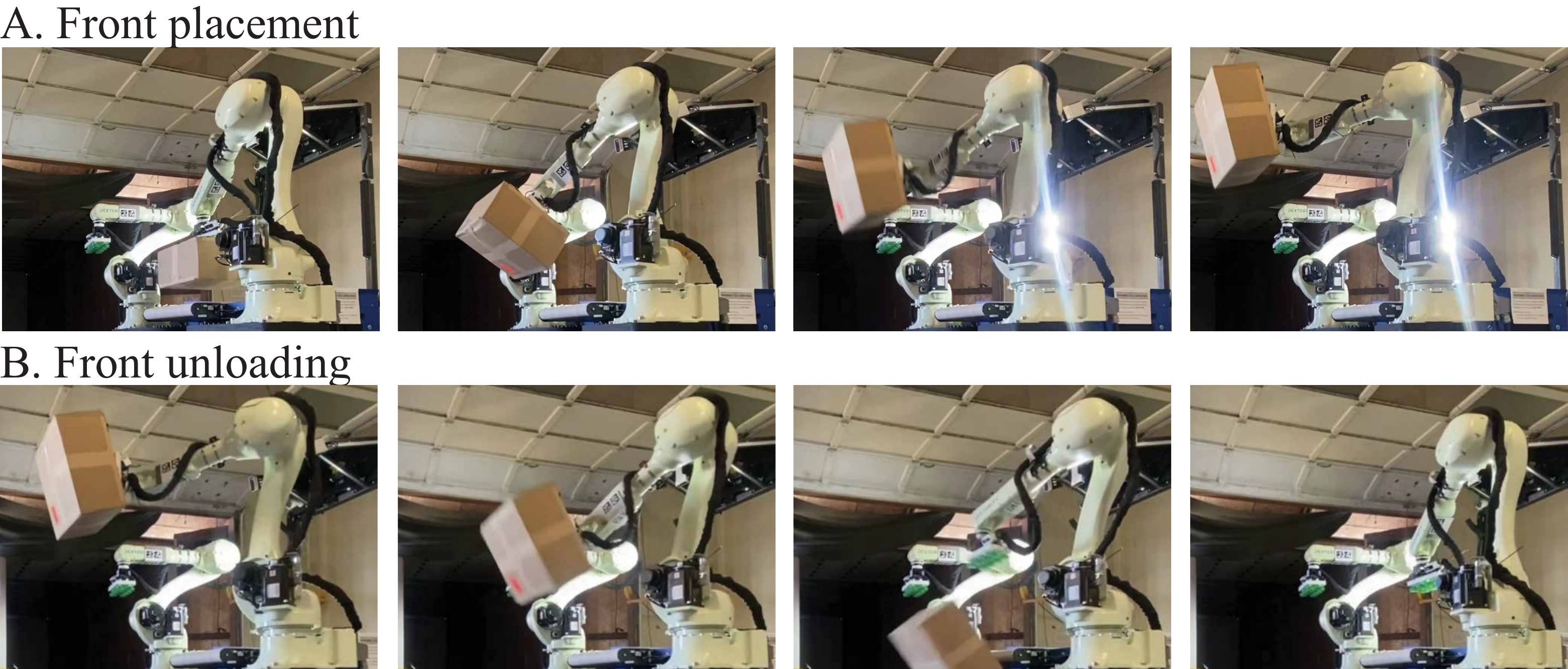}
    \vspace{-6mm}
    \caption{Test motion snapshots. A: Forward placement motion to move boxes from a conveyor to the top of a virtual stack of packages. B: Forward  unloading motions to return the box from the virtual stack to the conveyor.}
    \label{fig:snapshot}
    \vspace{-5mm}
\end{figure}

For practical evaluation, we extracted test motions from a real-world box-loading application. More concretely, we chose to perform front placement and front unloading motions, which required the robot to handle sideways grasps, increasing the risk of dropping or slipping the boxes, as described in Fig.~\ref{fig:snapshot}. Once the box was placed on the conveyor, the vision system estimated the box's dimensions and the robot picked it up. Then, an F/T sensor on the gripper estimated the box's weight. We used the estimated dimensions and weights of the boxes for the moment of inertia ($\mathbf{I}_o$) and mass ($m_o$) to compute grasp failure constraints. We assumed uniform density with $\mathbf{I}_o = \frac{1}{12}m~\text{diag}([b^2+c^2,c^2+a^2,a^2+b^2])$, where $[a,b,c]$ are their dimensions. However, without a sophisticated weight estimation method, the integrated system could only provide weight estimates with approximately $\pm 10 \%$ error. Additionally, if the mass of a box is concentrated at the bottom, the moment of inertia could increase by up to a factor of four.

To verify the algorithm accuracy, we conducted two experiments. The first one measured the success rate for pick-and-place maneuvers and the total motion duration with and without multi-suction-cup grasp failure constraints (MGFC). The second experiment involved reverse computing the maximum load that the robot could securely grasp based on the various grasp constraints. This was calculated for a set of trajectories that we had tested for front placement and front unloading of boxes of different sizes and weights. We recorded if the robot failed to maintain a grasp on the box throughout each trajectory. By comparing the actual box weights with the robot's grasp capability determined by grasp failure constraints, we evaluated the accuracy of each constraint by classifying the cases based on the following four criteria:
 

\begin{table}[h!] 
\label{tab:GraspCriterion}
\centering
\normalsize
\vspace{-2mm}
\begin{tabular}{l|l|l}
Prediction Accuracy & Grasp  & Estimated Max Load \\ \hline
True Positive (TP)  & Success & $\geq$ Actual Box Weight \\
True Negative (TN)  & Fail    & $\leq$ Actual Box Weight \\
False Negative (FN) & Success & $\leq$ Actual Box Weight \\
False Positive (FP) & Fail    & $\geq$ Actual Box Weight \\
\end{tabular}
\vspace{-2mm}
\end{table}

\subsubsection{analysis of results}

\begin{table}[!t]
\caption{motion duration[sec] and success rates[o/x] for \\ each trial motion without and with MGFC (multi-suction-cup grasp failure constraints) \label{tab:robot_result_1}}
\vspace{-2mm}
\centering
\begin{tabular}{c r r r}
\hline
trials $\#$ & without MGFC & with MGFC & time extension ($\%$) \\
\hline
\hline
front placement 1 & 1.259 (o) & 1.267 (o) & 0.6$\%$ \\
\rowcolor{LightCyan} front placement 2 & 1.215 (o) & 1.324 (o) & 8.23$\%$ \\
\rowcolor{LightCyan} front placement 3 & 1.207 (o) & 1.569 (o) & 19.2$\%$ \\
front placement 4 & 1.259 (o) & 1.259 (o) & 0$\%$ \\
\rowcolor{LightCyan} front placement 5 & 1.198 (o) & 1.378 (o) & 13.1$\%$ \\
front placement 6 & 1.197 (o) & 1.197 (o) & 0$\%$ \\
\rowcolor{LightRed} front placement 7 & 1.355 (x) & 2.038 (o) & 33.5$\%$ \\
front unloading 1 & 1.259 (o) & 1.290 (o) & 2.4$\%$ \\
front unloading 2 & 1.467 (o) & 1.494 (o) & 1.81$\%$ \\
\rowcolor{LightCyan} front unloading 3 & 1.215 (o) & 1.451 (o) & 16.2$\%$ \\
front unloading 4 & 1.259 (o) & 1.259 (o) & 0$\%$ \\
\rowcolor{LightRed} front unloading 5 & 1.201 (x) & 1.320 (o) & 9.02$\%$ \\
front unloading 6 & 1.197 (o) & 1.197 (o) & 0$\%$ \\
\rowcolor{LightRed} front unloading 7 & 1.482 (x) & 2.804 (o) & 89.2$\%$ \\
\hline
\end{tabular}
\vspace{-2mm}
\end{table}

\begin{table}[!t]
\setlength{\tabcolsep}{2pt}
\renewcommand{\arraystretch}{1.2}
\newcolumntype{a}{>{\columncolor{LightRed}}c}
\caption{Benchmark of Suction Grasp Failure Constraints Accuracy}
\vspace{-2mm}
\label{tab:robot_result_2}
\centering
\begin{tabular}{|l||c|c|c|a|}
\hline
 & True  & True  & False  & False  \\ Grasp Failure Constraints & Positive & Negative & Negative & Positive \\ \hline
w/ weak grasp stability \cite{pham2019critically} & 54$\%$ & 0$\%$ & 0$\%$ & 46$\%$ \\ \hline
w/ MGFC (w/o weight adj.)  & 59$\%$ & 18$\%$ & 12$\%$ & 12$\%$\\ \hline
w/ MGFC (w/ weight adj.) & 54$\%$ & 31$\%$ & 8$\%$ & 8$\%$\\ \hline
w/o grasp failure constraint & 22$\%$ & - & - & 78$\%$\\ \hline
\end{tabular}
\vspace{-3mm}
\end{table}

In Table~\ref{tab:robot_result_1}, the results of our first experiment are shown, measuring the motion duration and grasp success rate of front placements and front unloading motions with and without MGFC (multi-suction-cup grasp failure constraints). Without MGFC, the grasp failed in three instances (front placement 7, front unloading 5, front unloading 7). However, implementing MGFC also increased the motion duration by up to 19.2$\%$ (front placement 3), although the faster motion executed without MGFC was able to maintain the grasp throughout the trajectory. However, these unnecessary increases in motion duration are more likely caused by factors other than the inaccuracy of the proposed algorithm. The most significant one could be overestimated box weights. For instance, in front placement 3, where the actual box weight was \mbox{5.5 kg}, the estimated weight was \mbox{8.8 kg}, more than half heavier than the actual weight. Such inaccuracies in object information estimation are common in industrial settings, necessitating the use of safety factors.

Due to these complexities, conducting experiments in this manner may not fully demonstrate the load capacity of the trajectory, making it challenging to pinpoint factors contributing to longer motion durations or understand how trajectories without MGFC can maintain secure grasps. Instead, we provide a statistical analysis on how accurately the proposed MGFC prevents grasp failures compared to other methods, focusing on grasp failure prediction classification. The results are presented in Table~\ref{tab:robot_result_2}.

First, weak grasp stability~\cite{pham2019critically} overestimates load capacity predictions significantly, mostly exceeding the actual weight of boxes, even for trajectories where the grasp fails under the box's weight. Due to its optimism, which prevents it from predicting grasp failure, it struggles to accurately identify true negative cases. In contrast, our proposed algorithm demonstrates a more accurate approach. The proposed MGFC, using load distribution with weight adjustment, exhibits the lowest false positive (misses) rate compared to all other methods. While our algorithm shows clear improvements, occasional misses occur due to factors beyond our control, such as the actual mass distribution of a box and situations where a box is crumpled. Additionally, our algorithm reduces false negatives (false alarms) using proper weight parameters estimated for improved load distribution prediction.
 
 True negative ratio (TNR), calculated as ``(TN)/(TN+FP)" indicates how effectively the model identifies negative cases. We verified that the use of MGFC (w/ weight adj.) results in a higher TNR of 79$\%$,  than without MGFC (w/o weight adj.), which results on TNR of 60$\%$. Unfortunately, in our experimental setup, the weak grasp stability method never predicted a grasp failure due to its relaxed formulation of constraints, generating 0 $\%$ TNR. However, the false alarm rate (FAR = (FN)/(TP+FN)) is also higher for our proposed algorithm: 17$\%$ with MGFC (w/o weight adj.), with 13$\%$ for MGFC (W/ weight adj.), and 0$\%$ for weak grasp stability. In addition, it is important to note that a false alarm only leads to a slight increase in motion duration, which is preferable to missing a grasp in industrial applications.

%
%
\section{Conclusion}
\label{sec:conclusion}

In conclusion, this paper introduces an accurate analytical grasp failure model applicable to various configuration of vacuum grippers with multiple suction cups. Unlike using a single suction cup, predicting the load distributed on each suction cup exerted by the movement of an object during motion is crucial when employing a multi-suction-cup gripper.  We propose to solve this problem based on the principle of minimizing spring potential energy. We then verified that our load distribution model and grasp failure conditions closely match real data. For verification, we developed a testbed gripper equipped with force sensors capable of measuring load distribution across multiple suction cups. We also tested our model's ability to predict load distribution for various suction cup configurations and various objects to demonstrate the generalizability of the proposed model. 

Our proposed QP load distribution model offers an analytical solution, enabling a simple formulation for grasp failure constraints that can be easily integrated into trajectory optimization algorithms. Specifically, we demonstrated that our grasp failure model can be incorporated into time-optimal trajectory planning. While we have shown that our grasp failure model predicts grip failures well using a testbed gripper, we have substantiated its practical benefits on a real robot. We performed a comparative test on a real Kawasaki robot system running the same path with and without grasp failure constraints. The result showed that our model can effectively prevent the robot from incurring grasp failures.  

Additionally, to further validate the effectiveness of the proposed grasp failure constraints despite various uncertainties in real applications, such as inaccurate estimation of box weights and dimensions, we conducted a statistical accuracy test by comparing the actual load and maximum load capacity for several different grasp failure models. The results showed that our QP-based method offers a more accurate approach compared to the double description (DD) method. Although our grasp failure constraints may generate more false alarms, thereby limiting faster motion, our algorithm is significantly better at preventing grasp failures, which is our primary objective. While the algorithm demonstrates strong potential, it may not achieve perfect grasp failure prediction due to factors beyond our full control, such as internal mass movement within a box and potential box deformation during handling. In industry, it is common to apply a safety factor to prevent such failures, which can limit further the speed of the motions. In future research, we can explore enhancements to reduce both misses and false alarms, aiming to improve overall performance and efficiency.

\section*{Acknowledgments}
We thank Dexterity and the HCRL personnel for their support of this project, including Justin Kim, Jason Kmec, and Andrew Lovett for their work on the design and construction of the gripper testbed. Jee-eun Lee was a robotics intern for Dexterity, Inc. during the summer and fall of 2023 and the summer of 2024, and Luis Sentis was a consultant for Dexterity, Inc. during the summer of 2022.

{\appendix[Linear Programming formulation for Load distribution]
In the experimental analysis described in section~\ref{sec:experiment} B, we compared Linear Programming (LP) and Quadratic Programming (QP) methods to compute load distribution. In this section, we describe how we formulate LP load distribution. Considering the load distribution that minimizes the total sum of spring forces exerted at each suction cup, we have
\begin{align}
\text{min} ~& \| \overline{\mathbf{f}} \|_1 \label{eqn:LP_forcedistribution}\\
\textrm{subject to} ~& \mathcal{F}_t = \mathbf{A}\overline{\mathbf{f}} \nonumber
\end{align}
Here, $\|\cdot\|_1$ denotes the L1-norm, which is defined as $\|\mathbf{v}\|_1=\sum_{i=1}^{k} |v_i|$ for $\mathbf{v}=(v_1,\cdots,v_k)\in\mathbb{R}^k$.
To address absolute values in the optimization problem, we reformulate it by introducing the new optimization variable $\mathbf{x} = |\overline{\mathbf{f}}|$ as follows:
\begin{align*}
\min_{\overline{\mathbf{f}}, \mathbf{x}} ~& \mathbf{1}^\top \mathbf{x} \\
\textrm{subject to} ~& 
\begin{bmatrix}
    \mathbf{A} & 0
\end{bmatrix} 
\begin{bmatrix}
    \overline{\mathbf{f}} \\ \mathbf{x}
\end{bmatrix} = \mathcal{F}_t
\\
~& \begin{bmatrix}
    -\mathbf{I} & -\mathbf{I} \\ \mathbf{I} & -\mathbf{I}
\end{bmatrix} 
\begin{bmatrix}
    \overline{\mathbf{f}} \\ \mathbf{x}
\end{bmatrix} \leq 0.
\end{align*}
Finally, we obtain a load distribution problem formulated as an LP with linear equality and inequality constraints.
}


\bibliographystyle{IEEEtran}
\bibliography{bib/sample}

\begin{thebibliography}{10}
\providecommand{\url}[1]{#1}
\csname url@rmstyle\endcsname
\providecommand{\newblock}{\relax}
\providecommand{\bibinfo}[2]{#2}
\providecommand\BIBentrySTDinterwordspacing{\spaceskip=0pt\relax}
\providecommand\BIBentryALTinterwordstretchfactor{4}
\providecommand\BIBentryALTinterwordspacing{\spaceskip=\fontdimen2\font plus
\BIBentryALTinterwordstretchfactor\fontdimen3\font minus \fontdimen4\font\relax}
\providecommand\BIBforeignlanguage[2]{{%
\expandafter\ifx\csname l@#1\endcsname\relax
\typeout{** WARNING: IEEEtran.bst: No hyphenation pattern has been}%
\typeout{** loaded for the language `#1'. Using the pattern for}%
\typeout{** the default language instead.}%
\else
\language=\csname l@#1\endcsname
\fi
#2}}

\bibitem{reddy2013review}
P.~V.~P. Reddy and V.~Suresh, ``A review on importance of universal gripper in industrial robot applications,'' \emph{Int. J. Mech. Eng. Robot. Res}, vol.~2, no.~2, pp. 255--264, 2013.

\bibitem{jaiswal2017vacuum}
A.~Jaiswal and B.~Kumar, ``Vacuum cup grippers for material handling in industry,'' \emph{Int. J. Innov. Sci. Eng. Technol}, vol.~4, no.~6, pp. 187--194, 2017.

\bibitem{wirth2020suctionplate}
B.~Wirth, S.~Coutandin, and J.~Fleischer, ``Disturbance force estimation for a low pressure suction gripper based on differential pressure analysis,'' in \emph{Annals of Scientific Society for Assembly, Handling and Industrial Robotics}, 2020.

\bibitem{correll2016analysis}
N.~Correll, K.~E. Bekris, D.~Berenson, O.~Brock, A.~Causo, K.~Hauser, K.~Okada, A.~Rodriguez, J.~M. Romano, and P.~R. Wurman, ``Analysis and observations from the first amazon picking challenge,'' \emph{IEEE Transactions on Automation Science and Engineering}, vol.~15, no.~1, pp. 172--188, 2016.

\bibitem{mykhailyshyn2022gripping}
R.~Mykhailyshyn, V.~Savkiv, A.~M. Fey, and J.~Xiao, ``Gripping device for textile materials,'' \emph{IEEE Transactions on Automation Science and Engineering}, 2022.

\bibitem{zhang2020state}
B.~Zhang, Y.~Xie, J.~Zhou, K.~Wang, and Z.~Zhang, ``State-of-the-art robotic grippers, grasping and control strategies, as well as their applications in agricultural robots: A review,'' \emph{Computers and Electronics in Agriculture}, vol. 177, p. 105694, 2020.

\bibitem{jo2024suction}
Y.~Jo, Y.~Park, and H.~I. Son, ``A suction cup-based soft robotic gripper for cucumber harvesting: Design and validation,'' \emph{Biosystems Engineering}, vol. 238, pp. 143--156, 2024.

\bibitem{seretse2023material}
O.~Seretse, ``Material impact on performance of suction cups: A finite element analysis,'' \emph{J. Ind Intell}, vol.~1, no.~3, pp. 165--183, 2023.

\bibitem{joymungul2021gripe}
K.~Joymungul, Z.~Mitros, L.~Da~Cruz, C.~Bergeles, and S.~H. Sadati, ``Gripe-needle: a sticky suction cup gripper equipped needle for targeted therapeutics delivery,'' \emph{Frontiers in Robotics and AI}, vol.~8, p. 752290, 2021.

\bibitem{ichnowski2020gomp}
J.~Ichnowski, M.~Danielczuk, J.~Xu, V.~Satish, and K.~Goldberg, ``Gomp: Grasp-optimized motion planning for bin picking,'' in \emph{2020 IEEE International Conference on Robotics and Automation (ICRA)}.\hskip 1em plus 0.5em minus 0.4em\relax IEEE, 2020, pp. 5270--5277.

\bibitem{ichnowski2020deep}
J.~Ichnowski, Y.~Avigal, V.~Satish, and K.~Goldberg, ``Deep learning can accelerate grasp-optimized motion planning,'' \emph{Science Robotics}, vol.~5, no.~48, p. eabd7710, 2020.

\bibitem{ichnowski2022gomp}
J.~Ichnowski, Y.~Avigal, Y.~Liu, and K.~Goldberg, ``Gomp-fit: Grasp-optimized motion planning for fast inertial transport,'' in \emph{2022 International Conference on Robotics and Automation (ICRA)}.\hskip 1em plus 0.5em minus 0.4em\relax IEEE, 2022, pp. 5255--5261.

\bibitem{lynch1996dynamic}
K.~M. Lynch and M.~T. Mason, ``Dynamic underactuated nonprehensile manipulation,'' in \emph{Proceedings of IEEE/RSJ International Conference on Intelligent Robots and Systems. IROS'96}, vol.~2.\hskip 1em plus 0.5em minus 0.4em\relax IEEE, 1996, pp. 889--896.

\bibitem{lynch1999dynamic}
------, ``Dynamic nonprehensile manipulation: Controllability, planning, and experiments,'' \emph{The International Journal of Robotics Research}, vol.~18, no.~1, pp. 64--92, 1999.

\bibitem{zhu2002development}
J.~Zhu, D.~Sun, and S.-K. Tso, ``Development of a tracked climbing robot,'' \emph{Journal of Intelligent and robotic Systems}, vol.~35, pp. 427--443, 2002.

\bibitem{mantriota2007optimal}
G.~Mantriota, ``Optimal grasp of vacuum grippers with multiple suction cups,'' \emph{Mechanism and machine theory}, vol.~42, no.~1, pp. 18--33, 2007.

\bibitem{mantriota2007theoretical}
------, ``Theoretical model of the grasp with vacuum gripper,'' \emph{Mechanism and machine theory}, vol.~42, no.~1, pp. 2--17, 2007.

\bibitem{avigal2022gomp}
Y.~Avigal, J.~Ichnowski, M.~Y. Cao, and K.~Goldberg, ``Gomp-st: Grasp optimized motion planning for suction transport,'' in \emph{International Workshop on the Algorithmic Foundations of Robotics}.\hskip 1em plus 0.5em minus 0.4em\relax Springer, 2022, pp. 488--505.

\bibitem{de2018learning}
L.~De~Raedt, A.~Passerini, and S.~Teso, ``Learning constraints from examples,'' in \emph{Proceedings of the AAAI conference on artificial intelligence}, vol.~32, no.~1, 2018.

\bibitem{fajemisin2024optimization}
A.~O. Fajemisin, D.~Maragno, and D.~den Hertog, ``Optimization with constraint learning: A framework and survey,'' \emph{European Journal of Operational Research}, vol. 314, no.~1, pp. 1--14, 2024.

\bibitem{kleeberger2020survey}
K.~Kleeberger, R.~Bormann, W.~Kraus, and M.~F. Huber, ``A survey on learning-based robotic grasping,'' \emph{Current Robotics Reports}, vol.~1, pp. 239--249, 2020.

\bibitem{mahler2018dex}
J.~Mahler, M.~Matl, X.~Liu, A.~Li, D.~Gealy, and K.~Goldberg, ``Dex-net 3.0: Computing robust vacuum suction grasp targets in point clouds using a new analytic model and deep learning,'' in \emph{2018 IEEE International Conference on robotics and automation (ICRA)}.\hskip 1em plus 0.5em minus 0.4em\relax IEEE, 2018, pp. 5620--5627.

\bibitem{pham2019critically}
H.~Pham and Q.-C. Pham, ``Critically fast pick-and-place with suction cups,'' in \emph{2019 International Conference on Robotics and Automation (ICRA)}.\hskip 1em plus 0.5em minus 0.4em\relax IEEE, 2019, pp. 3045--3051.

\bibitem{caron2015leveraging}
S.~Caron, Q.-C. Pham, and Y.~Nakamura, ``Leveraging cone double description for multi-contact stability of humanoids with applications to statics and dynamics.'' in \emph{Robotics: science and systems}, vol.~11, 2015, pp. 1--9.

\bibitem{carpentier2018multicontact}
J.~Carpentier and N.~Mansard, ``Multicontact locomotion of legged robots,'' \emph{IEEE Transactions on Robotics}, vol.~34, no.~6, pp. 1441--1460, 2018.

\bibitem{fernbach2020c}
P.~Fernbach, S.~Tonneau, O.~Stasse, J.~Carpentier, and M.~Ta{\"\i}x, ``C-croc: Continuous and convex resolution of centroidal dynamic trajectories for legged robots in multicontact scenarios,'' \emph{IEEE Transactions on Robotics}, vol.~36, no.~3, pp. 676--691, 2020.

\bibitem{lee2022adaptive}
J.~Lee, T.~Bandyopadhyay, and L.~Sentis, ``Adaptive robot climbing with magnetic feet in unknown slippery structure,'' \emph{Frontiers in Robotics and AI}, vol.~9, p. 949460, 2022.

\bibitem{lynch2017modern}
K.~M. Lynch and F.~C. Park, \emph{Modern robotics}.\hskip 1em plus 0.5em minus 0.4em\relax Cambridge University Press, 2017.

\bibitem{caron2015stability}
S.~Caron, Q.-C. Pham, and Y.~Nakamura, ``Stability of surface contacts for humanoid robots: Closed-form formulae of the contact wrench cone for rectangular support areas,'' in \emph{2015 IEEE International Conference on Robotics and Automation (ICRA)}.\hskip 1em plus 0.5em minus 0.4em\relax IEEE, 2015, pp. 5107--5112.

\bibitem{chettibi2004minimum}
T.~Chettibi, H.~Lehtihet, M.~Haddad, and S.~Hanchi, ``Minimum cost trajectory planning for industrial robots,'' \emph{European Journal of Mechanics-A/Solids}, vol.~23, no.~4, pp. 703--715, 2004.

\bibitem{diehl2006fast}
M.~Diehl, H.~G. Bock, H.~Diedam, and P.-B. Wieber, ``Fast direct multiple shooting algorithms for optimal robot control,'' \emph{Fast Motions in Biomechanics and Robotics}, pp. 65--93, 2006.

\bibitem{schulman2013finding}
J.~Schulman, J.~Ho, A.~X. Lee, I.~Awwal, H.~Bradlow, and P.~Abbeel, ``Finding locally optimal, collision-free trajectories with sequential convex optimization.'' in \emph{Robotics: Science and Systems}, vol.~9, no.~1.\hskip 1em plus 0.5em minus 0.4em\relax Berlin, Germany, 2013.

\bibitem{zhao2018efficient}
Y.~Zhao, H.-C. Lin, and M.~Tomizuka, ``Efficient trajectory optimization for robot motion planning,'' in \emph{International Conference on Control, Automation, Robotics and Vision (ICARCV)}.\hskip 1em plus 0.5em minus 0.4em\relax IEEE, 2018, pp. 260--265.

\bibitem{zhang2022time}
X.~Zhang, F.~Xiao, X.~Tong, J.~Yun, Y.~Liu, Y.~Sun, B.~Tao, J.~Kong, M.~Xu, and B.~Chen, ``Time optimal trajectory planning based on improved sparrow search algorithm,'' \emph{Frontiers in Bioengineering and Biotechnology}, vol.~10, 2022.

\bibitem{wen2022path}
Y.~Wen and P.~Pagilla, ``Path-constrained and collision-free optimal trajectory planning for robot manipulators,'' \emph{IEEE Transactions on Automation Science and Engineering}, vol.~20, no.~2, pp. 763--774, 2022.

\bibitem{shin1985minimum}
K.~Shin and N.~McKay, ``Minimum-time control of robotic manipulators with geometric path constraints,'' \emph{IEEE Transactions on Automatic Control}, vol.~30, no.~6, pp. 531--541, 1985.

\bibitem{shiller1989robot}
Z.~Shiller and S.~Dubowsky, ``Robot path planning with obstacles, actuator, gripper, and payload constraints,'' \emph{The International Journal of Robotics Research}, vol.~8, no.~6, pp. 3--18, 1989.

\bibitem{shiller1991computing}
------, ``On computing the global time-optimal motions of robotic manipulators in the presence of obstacles,'' \emph{IEEE Transactions on Robotics and Automation}, vol.~7, no.~6, pp. 785--797, 1991.

\bibitem{kavraki1996probabilistic}
L.~E. Kavraki, P.~Svestka, J.-C. Latombe, and M.~H. Overmars, ``Probabilistic roadmaps for path planning in high-dimensional configuration spaces,'' \emph{IEEE transactions on Robotics and Automation}, vol.~12, no.~4, pp. 566--580, 1996.

\bibitem{lavalle1998rapidly}
S.~LaValle, ``Rapidly-exploring random trees: A new tool for path planning,'' \emph{Research Report 9811}, 1998.

\bibitem{kuffner2000rrt}
J.~J. Kuffner and S.~M. LaValle, ``Rrt-connect: An efficient approach to single-query path planning,'' in \emph{Proceedings 2000 ICRA. Millennium Conference. IEEE International Conference on Robotics and Automation. Symposia Proceedings (Cat. No. 00CH37065)}, vol.~2.\hskip 1em plus 0.5em minus 0.4em\relax IEEE, 2000, pp. 995--1001.

\bibitem{karaman2011sampling}
S.~Karaman and E.~Frazzoli, ``Sampling-based algorithms for optimal motion planning,'' \emph{The international journal of robotics research}, vol.~30, no.~7, pp. 846--894, 2011.

\bibitem{devaurs2015optimal}
D.~Devaurs, T.~Sim{\'e}on, and J.~Cort{\'e}s, ``Optimal path planning in complex cost spaces with sampling-based algorithms,'' \emph{IEEE Transactions on Automation Science and Engineering}, vol.~13, no.~2, pp. 415--424, 2015.

\bibitem{gammell2015batch}
J.~D. Gammell, S.~S. Srinivasa, and T.~D. Barfoot, ``Batch informed trees (bit*): Sampling-based optimal planning via the heuristically guided search of implicit random geometric graphs,'' in \emph{2015 IEEE international conference on robotics and automation (ICRA)}.\hskip 1em plus 0.5em minus 0.4em\relax IEEE, 2015, pp. 3067--3074.

\bibitem{pfeiffer1987concept}
F.~Pfeiffer and R.~Johanni, ``A concept for manipulator trajectory planning,'' \emph{IEEE Journal on Robotics and Automation}, vol.~3, no.~2, pp. 115--123, 1987.

\bibitem{verscheure2009time}
D.~Verscheure, B.~Demeulenaere, J.~Swevers, J.~De~Schutter, and M.~Diehl, ``Time-optimal path tracking for robots: A convex optimization approach,'' \emph{IEEE Transactions on Automatic Control}, vol.~54, no.~10, pp. 2318--2327, 2009.

\bibitem{pham2018new}
H.~Pham and Q.-C. Pham, ``A new approach to time-optimal path parameterization based on reachability analysis,'' \emph{IEEE Transactions on Robotics}, vol.~34, no.~3, pp. 645--659, 2018.

\bibitem{lee2024performance}
J.~Lee, A.~Bylard, R.~Sun, and L.~Sentis, ``On the performance of jerk-constrained time-optimal trajectory planning for industrial manipulators,'' \emph{arXiv preprint arXiv:2404.07889}, 2024.

\bibitem{golberg1989genetic}
D.~E. Golberg, ``Genetic algorithms in search, optimization, and machine learning,'' \emph{Addion wesley}, vol. 1989, no. 102, p.~36, 1989.

\bibitem{conn1991globally}
A.~R. Conn, N.~I. Gould, and P.~Toint, ``A globally convergent augmented lagrangian algorithm for optimization with general constraints and simple bounds,'' \emph{SIAM Journal on Numerical Analysis}, vol.~28, no.~2, pp. 545--572, 1991.

\bibitem{conn1997globally}
A.~Conn, N.~Gould, and P.~Toint, ``A globally convergent lagrangian barrier algorithm for optimization with general inequality constraints and simple bounds,'' \emph{Mathematics of computation}, vol.~66, no. 217, pp. 261--288, 1997.

\bibitem{GlobalOptimizationToolbox}
\BIBentryALTinterwordspacing
T.~M. Inc., ``Global optimization toolbox version: 24.1.0 (r2024a),'' Natick, Massachusetts, United States, 2024. [Online]. Available: \url{https://www.mathworks.com}
\BIBentrySTDinterwordspacing

\end{thebibliography}

\end{document}